%% file: main.tex
\documentclass[review]{elsarticle}

\journal{Journal of \LaTeX\ Templates}

\usepackage{amssymb}
\usepackage{graphicx}
\usepackage{booktabs}
\usepackage{color}
\usepackage[ruled,linesnumbered,boxed]{algorithm2e}
\usepackage{mathtools}
\usepackage{soul}
\usepackage{diagbox}
\usepackage{subcaption}
\usepackage{caption}
\usepackage{pbox}
\usepackage{float}
\usepackage{mathtools}
\SetKwComment{Comment}{$\triangleright$\ }{}

\SetKwProg{Dopar}{do parallel}{}{end}
\SetKwInput{kwInput}{Input}
\SetKwInput{kwOutput}{Output}
\usepackage{lscape} 
\SetKwRepeat{Do}{do}{while}

\usepackage{url}
\urldef{\mailsa}\path|{blaz.skrlj,jan.kralj,nada.lavrac}@ijs.si|
\usepackage{siunitx}
\usepackage{amsmath}

\SetKwInOut{Parameter}{Parameters}
\newcommand{\finaldatanum}{15 }
\newcommand{\finalsynthetic}{1488 }

\usepackage{graphicx}
\begin{document}

\begin{frontmatter}

\title{Deep Node Ranking for Neuro-symbolic Structural Node Embedding and Classification}
%\title{Deep Node Ranking: Structural Embedding and End-to-End Network Node Classification}

%% Group authors per affiliation:
\author{Bla\v{z} \v{S}krlj}
\address{Jo\v{z}ef Stefan Institute, Jamova 39, 1000 Ljubljana, Slovenia \\Jo\v{z}ef Stefan Int. Postgraduate School, Jamova 39, 1000 Ljubljana, Slovenia}
\fntext[myfootnote]{blaz.skrlj@ijs.si}

\author{Jan Kralj}
\address{Jo\v{z}ef Stefan Institute, Jamova 39, 1000 Ljubljana, Slovenia \\
Cosylab d.o.o., Ljubljana, Slovenia}

\author{Janez Konc}
\address{National Institute of Chemistry, Ljubljana, Slovenia}

\author{Marko Robnik-\v{Sikonja}}
\address{Faculty of Computer and Information Science, Ljubljana, Slovenia}

\author{Nada Lavra\v{c}}
\address{Jo\v{z}ef Stefan Institute, Jamova 39, 1000 Ljubljana, Slovenia \\ University of Nova Gorica, Vipavska 13, 5000 Nova Gorica, Slovenia}

\begin{abstract}
Network node embedding is an active research subfield of complex network analysis. This paper contributes a novel approach to learning network node embeddings and direct node classification using a node ranking scheme coupled with an autoencoder-based neural network architecture. The main advantages of the proposed Deep Node Ranking (DNR) algorithm are competitive or better classification performance, significantly higher learning speed and lower space requirements when compared to state-of-the-art approaches on \finaldatanum real-life node classification benchmarks. Furthermore, it enables exploration of the relationship between symbolic and the derived sub-symbolic node representations, offering insights into the learned node space structure.
To avoid the space complexity bottleneck in a direct node classification setting, DNR computes stationary distributions of personalized random walks from given nodes in mini-batches, scaling seamlessly to larger networks. The scaling laws associated with DNR were also investigated on \finalsynthetic  synthetic Erd\H{o}s-R\'enyi networks, demonstrating its scalability to tens of millions of links.
\end{abstract}


\end{frontmatter}
%\linenumbers

\section{Introduction}
Numerous real-world systems consisting of interconnected entities can be represented as complex networks. Analysis of such networks provides insights into the underlying patterns applicable 
in various practical scenarios, including the discovery of drug targets, modelling of disease outbreaks, author profiling, modelling of transportation and the study of social dynamics~\cite{benson2016higher}.

Modern machine learning approaches applied to complex networks offer intriguing opportunities for developing fast and accurate algorithms that can learn based on the structural topology of a given network. Recently, approaches based on network node embedding \citep{grover2016node2vec,zitnik2017predicting,struc2vec} became prevalent for many common tasks, such as node classification, edge prediction and unsupervised node clustering (community detection).
Node embedding refers to the process of learning node representations in a numeric vector format that captures the topological properties of network nodes~\cite{perozzi2014deepwalk}. 
Embeddings are useful, as vector {representations} are suitable for conventional machine learning algorithms capable of {addressing the tasks from classification and regression to clustering}.

In this work, we propose a new network {node embedding} and classification algorithm named \emph{Deep Node Ranking (DNR)}, which combines efficient node ranking with the non-linear approximation power of deep neural networks. The developed framework uses deep neural networks to obtain {node} embeddings directly from \emph{stationary random walk distributions} produced by random walkers with a restart with respect to individual nodes of interest. {Compared to existing methods, DNR is, to our knowledge, one of the first \emph{neuro-symbolic} node representation learning algorithms, as it offers joint construction of low-dimensional latent representations via symbolic (inspectable) node features.}

Even though there already exist embedding approaches based on higher-order random walks \citep{grover2016node2vec, struc2vec} (i.e. random walkers with memory), the stationary distribution of first-order random walkers {has not yet been fully explored in a deep learning setting}. {Widely used methods such as node2vec and struc2vec perform well; however, they do not necessarily scale to larger networks and often require extensive hyperparameter tuning for good performance. Further, these methods mostly learn representations via rather shallow, single latent layer-like optimization schemes, potentially missing the abstraction learning power of deeper neural networks. Finally, in massive networks, not all nodes need to be accounted for during representation learning -- information relevant to representing a given node can depend on its relation to a small number of key nodes. We demonstrate that the neuro-symbolic paradigm offers an elegant solution to this problem via ranking-based pivoting (selection of symbolic features before deep learning), scaling to networks comprised of tens of millions of links and tens of thousands of nodes on commodity hardware. This paper also offers ablation studies of DNR's scalability on more than 1{,}400 synthetic networks of different sizes -- this type of analysis is seldom considered in related work.}

We showcase the developed algorithm's capabilities on the challenging problems of node classification and network visualization, highlighting its ability to learn and accurately predict node labels {at scale}. Further, we compiled one of the largest collections of node classification data sets and used it for empirical evaluation of the methods. Key contributions of this paper are:
\begin{enumerate}
\item A fast network embedding algorithm named Deep Node Ranking (DNR) based on global personalized node ranks. It performs competitively and can be used for a multitude of downstream learning tasks, including node classification, network visualization and similar. The proposed neuro-symbolic algorithm is also faster than many state-of-the-art embedding algorithms, {and scales better}.
\item To our knowledge, 
{the proposed node embedding algorithms are for the first time benchmarked against contemporary approaches on such scale} (\finaldatanum real data sets), as commonly the algorithms are tested only on a handful of data sets.
\item {We conducted an extensive empirical evaluation on} \finalsynthetic {synthetic networks to study the effects of node pivoting, for which we hypothesized that it substantially improves scalability.}
\item {The proposed DNR performs better than the competition when the labelled data is scarce (small percentage of labelled nodes).}
\end{enumerate}

The remainder of this work is structured as follows. In Section \ref{sec:related}, we shortly review the related work on neuro-symbolic representation learning, network node classification and network node ranking. Section \ref{sec:DNR_origin} presents the proposed network node embedding algorithm that combines deep neural networks with network node ranking. In Section \ref{sec:experimental}, we describe the experimental setting and different non-synthetic complex networks from different domains used in the evaluation, including the newly composed data sets. The experimental results are presented in Section \ref{sec:results}. In Section \ref{sec:conclusions} we conclude the work and present plans for further work.

\section{Background and related work}
\label{sec:related}

This section presents deep and neuro-symbolic learning preliminaries that describe how algorithms learn from complex networks and what is learned, followed by an overview of node ranking algorithms relevant to this work.

\subsection{Neuro-symbolic representation learning}
\label{sec:nsl}
{We first discuss the branch of methods that exploits the insights from the fields of deep learning and symbolic learning, which can be referred to as \emph{neuro-symbolic representation learning}.
This paradigm of learning has been actively studied for the past twenty years (see~\cite{garcez2020neurosymbolic}); however, it resurged recently with many works that demonstrated this paradigm's utility when compared to symbolic/sub-symbolic-only learning.
The interest in neuro-symbolic learning, for example, spiked recently {\cite{Mao2019NeuroSymbolic}} by the development of a neuro-symbolic system that partially operates via symbolic and partially via a sub-symbolic space, used to distil human-understandable concepts from images. The recent work on closing the loop between recognition (neural) and reasoning (symbolic) {\cite{li2020ngs}} introduced a grammar model as a symbolic prior to bridge neural perception and symbolic reasoning, alongside a top-down, human-like induction procedure. This work demonstrated that such a combined approach significantly outperforms the conventional reinforcement learning-based baselines. The Microsoft research division (MSR) recently explored the interplay between visual recognition and reasoning {\cite{amizadeh2020neuro-symbolic}}. They introduced a framework to isolate and evaluate the reasoning aspect of visual question answering separately from its perception, followed by a calibration procedure that offers an exploration of the relation between reasoning and perception. Further, a neuro-symbolic approach to logical deduction was proposed as \emph{Neural Logic Machines} {\cite{nlogm}}. This architecture was shown to have inductive logic learning capabilities, which was demonstrated on simple tasks such as sorting. Finally, the two recent approaches from the field of inductive logic programming (ILP) explored the interplay between the logical input structures and how they perform when associated with neural network-based learning. The Deep Relational Machines{~\cite{DRM}} were one of the first approaches to showcase the utility of combining the two paradigms. Further, the recent work of Srinivasan et al.~{\cite{JMLR:v20:18-517}} explored how Deep Relational Machines can be \emph{explained}, emphasizing that being able to explain what a given association system does is highly relevant in e.g., the field of biomedicine.

{In the last years, links between sub-symbolic and logic programming were also established. For example, the DeepProbLog system~{\cite{manhaeve2018deepproblog}} demonstrates how neural predicates could be useful for constructing expressive (and short) programs for complex tasks such as image-based enumeration. Furthermore, links between statistical learning and the neuro-symbolic paradigm were also studied~{\cite{deraedt2020statistical}}. Finally, recent endeavours in this direction also introduce the notion of stochasticity as a programming component~{\cite{winters2021deepstochlog}}.}

Albeit being actively explored, the notion of neuro-symbolic representation learning was, to our knowledge, not yet considered in the context of node representation learning, which is the key focus of this work.
}

\subsection{Network node classification}
\label{sec:learning_cn}
 Complex networks, representing real-world phenomena such as financial markets, transportation, biological interactions or social dynamics \cite{benson2016higher,nowzari2016analysis,le2015novel} often possess interesting properties such as scale invariance,  non-trivial partitioning, presence of hub nodes, weakly connected components, heavy-tailed node degree distributions, occurrence of communities, significant motif pattern counts, etc. ~\cite{costa2007characterization,van2016random}. \emph{Learning from complex networks} considers different aspects of complex networks, e.g., network structure and node labels, which are used as inputs to machine learning algorithms to address learning tasks such as link prediction, node classification, etc.

In this paper we focus on node classification, i.e. the problem of classifying nodes into {two or more} distinct classes. {This task is considered as semi-supervised learning}, given that the whole network is used to obtain representations of individual nodes, from which the network classification model is learned.
Information propagation algorithms~\cite{Zhu02learningfrom} propagate label information via nodes' neighbors until all nodes are labeled. These algorithms learn in an \emph{end-to-end} manner, meaning that no intermediary representation of a network is first obtained and subsequently used for training e.g., a classifier.

Another class of node classification algorithms learns node labels from node embeddings, i.e. node representations in vector form~\cite{cui2018survey}. Here, the whole network is first transformed into an information-rich, {compact} low-dimensional representation {(a dense matrix)}. This representation serves as an input to plethora of more general machine learning approaches that can be used for node classification.

We distinguish between two main branches of embedding-based learning algorithms, discussed next: graph neural networks and random walk-based learners. Graph Neural Networks (GNNs), introduced in the recent years, attempt to incorporate a given network's adjacency structure as {new} neural network layers. Amongst first such approaches were the Graph Convolutional Networks (GCNs)~\cite{kipf2016semi}, {their} generalization with the attention mechanism \cite{velivckovic2018graph}, {and} the more recent isomorphism-based variants with provable properties~\cite{xu2018powerful}. Treating the adjacency structure as a neural network has also shown promising results~\cite{hamilton2017inductive}. The key characteristic of this branch of methods is their capability {to account for} \emph{node features} by multiplication of the normalized adjacency matrix as part of a special layer during learning from features. On the other hand, if node features are not available, which is the case with the majority of freely available public data sets, more optimized methods focused on \emph{structure-based learning} {are preferred}.  For example, the LINE algorithm~\cite{tang2015line} uses the network's \emph{eigendecomposition} in order to learn a low dimensional network representation, e.g., a representation of the network{'s nodes} in 128 dimensions instead of the dimension {that} matches the number of nodes. Approaches that use random walks to sample the network include DeepWalk \cite{perozzi2014deepwalk} and its generalization node2vec \cite{grover2016node2vec}. It was recently proven that DeepWalk, node2vec, and LINE can be reformulated as implicit matrix factorization \cite{Qiu:2018:NEM:3159652.3159706}. Furthermore, approaches such as struc2vec~\cite{struc2vec} demonstrated how more complex, multilayer structure can be compressed into node representations for better performance.
Despite many promising approaches developed, a recent extensive evaluation of network embedding techniques \cite{goyal2017graph} suggests that node2vec \cite{grover2016node2vec} remains one of the best embedding approaches for the task of \textbf{structural} node classification.

\subsection{Network node ranking}
\label{sec:nr}

Node ranking algorithms assess the relevance of a node in a network
either globally (relative to the whole network) or locally (relative to a sub-network) by assigning a \emph{score} (i.e. \emph{rank}) to each node in the network. In this work we only consider node ranking algorithms that compute a local relevance score of a node based on its direct neighborhood.
The key such node ranking algorithm is the Personalized PageRank (P-PR) algorithm \citep{pagerank}, sometimes referred to as random walk with restart \citep{tong2006fast}. Personalized PageRank uses random walks to calculate the relevance of nodes in a network. It obtains the stationary distribution of a random walk that starts at {a given node}. The P-PR-based approaches were used successfully to study cellular networks, social phenomena \cite{halu2013multiplex}, and many other real-world networks \cite{yu2017pagerank}. Efficient implementation of P-PR algorithms remains an active research field, for example, the recent bidirectional variation of the P-PR was introduced {to speed up} the node ranking process \cite{lofgren2016personalized}.
The obtained stationary distribution of a random walk can be used directly for network-based learning tasks, as demonstrated in HINMINE methodology \cite{kralj2017hinmine}. 

\subsection{Combining node ranking and node representation learning}
{Exploring} the ideas of augmenting learning with ranking was in the recent years explored in the context of graph neural networks. For example, ranking was used to prioritize propagation \cite{klicpera2018predict} {and} to scale graph neural networks \cite{bojchevski2019pagerank}. A similar idea was exploited by~\cite{xu2018representation}, where a more efficient propagation scheme was proposed {by using} node ranking. The proposed Deep Node Ranking algorithm is novel with respect to these works, as it {exploits} both {the} fast, parallel personalized node rank computation and the \emph{representation learning} power of deep neural networks.

\section{Deep Node Ranking}
\label{sec:DNR_origin}

\begin{figure}[h!]
\centering
\includegraphics[width=0.85\linewidth]{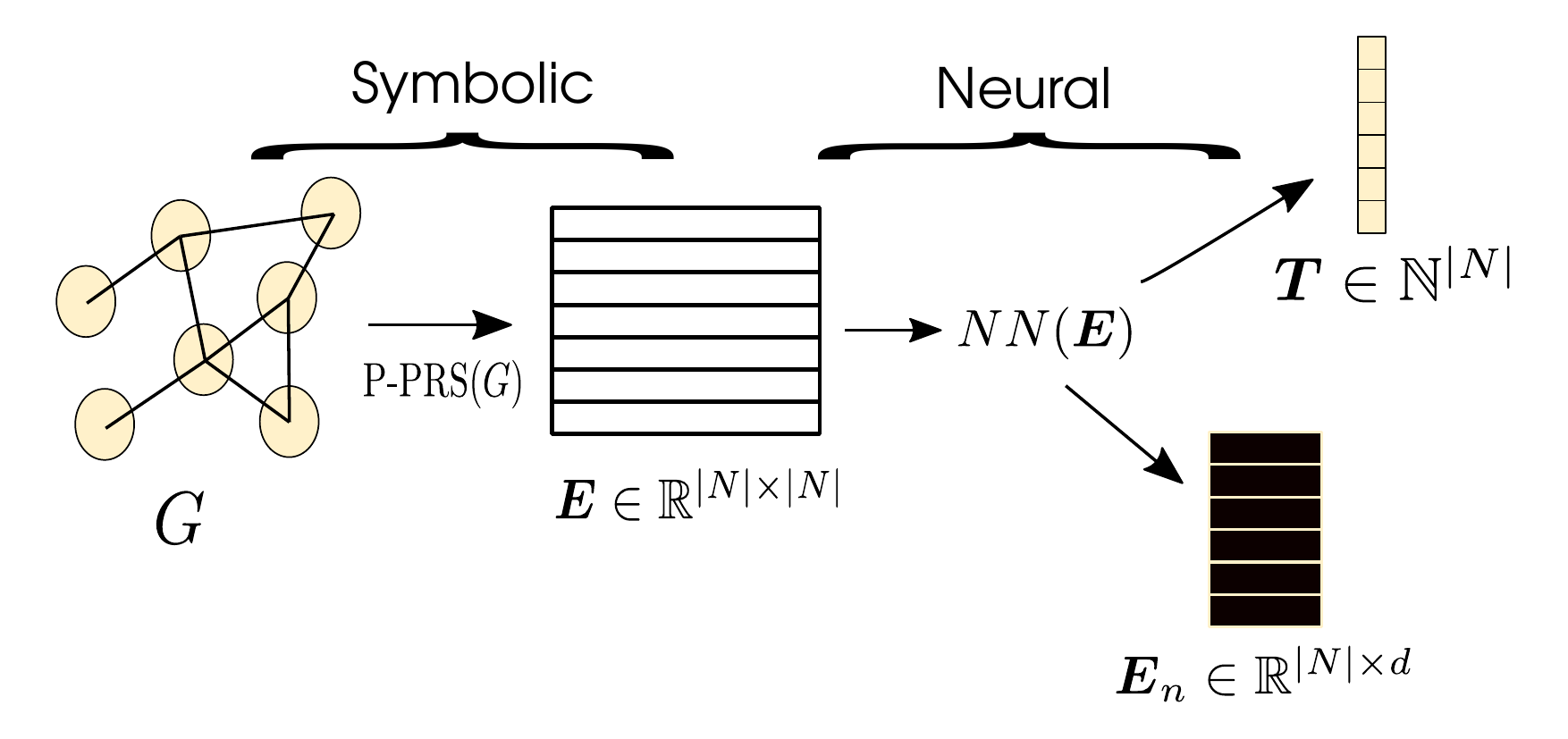}
\caption{Deep Node Ranking algorithm. The symbolic part of the algorithm computes personalized PageRank vectors ($\boldsymbol{E}$), which are subsequently compressed with a neural network ($\textrm{NN}$) into either a lower dimensional representation ($\boldsymbol{E}_n$), or used for end-to-end classification ($\boldsymbol{T}$). Note that the intermediary symbolic \textsc{P-PRS} based representation remains interpretable (features are nodes).}
\label{scheme}
\end{figure}

This section presents the Deep Node Ranking (DNR) algorithm for neuro-symbolic structural network node embedding and end-to-end node classification (overview shown in Figure \ref{scheme}). The name of the algorithm, Deep Node Ranking, reflects the two main ideas considered: network node ranking step (\emph{symbolic}) and the subsequent deep neural network learning step (\emph{neural}/\emph{sub-symbolic}).
In the first step of DNR, personalized node ranks are computed for each node, resulting in Personalized PageRank with shrinking (P-PRS) vectors. {These vectors are symbolic, as each dimension corresponds to a given node.}
In the second step, the P-PRS vectors are considered by a deep neural network consisting of at least a single dense embedding layer {of} size equal to the predefined embedding dimension.
{This embedding is sub-symbolic, as one can no longer interpret the meaning of individual (latent) dimensions.}
The third, output step, consists either of an output layer with the number of its neurons equal to the number of target classes (top) enabling direct classification of nodes or embeddings (bottom), which correspond to the embedding layer from Step 2. The {obtained} embeddings can be used for downstream machine learning tasks, such as classification, network visualization, and comparison.

The DNR algorithm, which takes as input a partially labeled complex network, consists of three steps {outlined below}.
\begin{enumerate}
    \item Network node ranking results in learned node representations, obtained by using the Personalized PageRank with Shrinking (P-PRS) algorithm. This step results in a matrix of P-PRS vectors of dimension $|N|$.
	\item Representation distillation. A neural network architecture compresses the prepared personalized {PageRank} vectors into compact representations (of dimension $d$).
   \item Output phase. The output of the network can be either node classification, that is, direct learning of node labels or a collection of low-dimensional node representations.
\end{enumerate}

\subsection{Node ranking with the Personalized PageRank with Shrinking algorithm}
\label{sec:PPR}

We first build and upgrade the representation learning idea, introduced in previous work \citep{kralj2017hinmine}, where node representations are obtained via personalized node ranking. {The following description represents a substantial theoretical extension of the original idea, which was further parallelized for the first time in this work. Furthermore, this work introduces node pivoting, which substantially speeds up the personalized ranking time.}
We consider a version of the Personalized PageRank \cite{pagerank1999} algorithm to which we refer to as Personalized PageRank with Shrinking ($\textsc{P-PRS}$) (algorithm~\ref{algo:PPR}). This variant of the widely known PPR algorithm produces node representations (or $\textsc{P-PRS}$ vectors) by simulating random walks for each node of the input network. Compared to the network adjacency matrix, $\textsc{P-PRS}$  vectors contain traversal information for each node, reflecting its ranking based on a node's position with respect to a given network's topology.  \\

\begin{algorithm}[H]
\scriptsize
\KwData{A complex network's adjacency matrix $A$, with nodes $N$ and edges $E$, starting node $u \in N$}
\Parameter{damping factor $\delta$, spread step $\sigma$, spread percent $\tau$ (default 50\%), stopping criterion $\epsilon$}
\KwResult{$\textsc{P-PRS}_u$ vector describing stationary distribution of random walker visits with respect to $u \in N$}
$A$ $\leftarrow$ toRightStochasticMatrix($A$)\Comment*[r]{Transpose and normalize rows of $A$}
core\_vector $\leftarrow$ $[0,\dots,0] $\Comment*[r]{Initialize zero vector of size |N|}
core\_vector$[u]$ $\leftarrow$ $1$;
rank\_vector $\leftarrow$ $\textrm{core\_vector}$;
$v$ $\leftarrow$ $\textrm{core\_vector}$\;
$\textrm{steps}$ $\leftarrow$ 0\Comment*[r]{Shrinking part}
nz $\leftarrow$ 1\Comment*[r]{Number of non-zero P-PRS values}
\While{$\textrm{nz} < |N| \cdot \tau$ $\wedge$ $\textrm{steps} < \sigma$}{ 
  $\textrm{steps} \leftarrow \textrm{steps} + 1$\;
  $v = v + A \cdot v$\Comment*[r]{Update transition vector}
  $\textrm{nzn}$ $\leftarrow$ $\textrm{nonZero}(v)$\Comment*[r]{Identify non-zero values}
  \If{$\textrm{nzn} = \textrm{nz}$}{
    $\textrm{shrink}$ $\leftarrow$ $\textrm{True}$\;
    \textbf{end while}\;
  }
  $\textrm{nz}$ $\leftarrow$ $\textrm{nzn}$\;
}
\If{$\textrm{shrink}$}{
  $\textrm{toReduce}$ $\leftarrow$ $\{i; v[i]\neq0\}$\Comment*[r]{Indices of non-zero entries in vector $v$}
  core\_rank $\leftarrow$ $\textrm{core\_rank}[\textrm{toReduce}]$;
  rank\_vector $\leftarrow$ $\textrm{rank\_vector}[\textrm{toReduce}]$\;
  $A$ $\leftarrow$ $A[\textrm{toReduce},\textrm{toReduce}]$\Comment*[r]{Shrink  a sparse adjacency matrix}
}
$diff$ $\leftarrow$ $\infty$\;
$steps$ $\leftarrow$ 0\Comment*[r]{Node ranking - power iteration}
\While{$ \textrm{diff} > \epsilon \wedge steps < max\_steps$}{
  $\textrm{steps} \leftarrow \textrm{steps} + 1$\;
  new\_rank $\leftarrow$ $A \cdot \textrm{rank\_vector}$;
  rank\_sum $\leftarrow$ $\sum_{i}\textrm{rank\_vector}[i]$\;
  \If{$rank\_sum < 1 $}{ 
    new\_rank $\leftarrow$ new\_rank + start\_rank $\cdot$ $(1 - \textrm{rank\_sum})$\;
  }
  new\_rank $\leftarrow$ $\delta \cdot \textrm{new\_rank} + (1 - \delta) \cdot \textrm{start\_rank}$\;
  \textrm{diff} $\leftarrow$ $\|\textrm{rank\_vec}-\textrm{new\_rank}\|$\Comment*[r]{Norm computation}
  rank\_vec $\leftarrow$ new\_rank\;    
} 
\uIf{$shrink$}{
  $\textsc{P-PRS}_u$ $\leftarrow$ $[0,\dots,0]$\Comment*[r]{Zero vector of dimension |N|}
  $\textsc{P-PRS}_u[\textrm{toReduce}]$ $\leftarrow$ rank\_vec\;
}
\uElse{$\textsc{P-PRS}_u$ $\leftarrow$ rank\_vec}
\Return $\textsc{P-PRS}_u$\;
\caption{P-PRS: Personalized PageRank with Shrinking}
 \label{algo:PPR}
\end{algorithm}
The \textsc{P-PRS} algorithm consists of two main parts:
\begin{enumerate}
\item 
In the first part named the \emph{shrinking step} (lines 5--20 of Algorithm~\ref{algo:PPR}), in each iteration, the walker spreads from nodes with non-zero PageRank values to their neighbors. 
\item In the second part of the algorithm, named the \emph{P-PRS computation step} (lines 23--38 of Algorithm~\ref{algo:PPR}), P-PRS vectors corresponding to individual network nodes are computed using the power iteration method (Eq.~\ref{eqPR}).
\end{enumerate}

\noindent {\bf Shrinking step.} In the shrinking step we take into account the following: 
\begin{itemize}
\item If no path exists between node $u$ (the starting node) and node $i$, the $\textsc{P-PRS}$ value assigned to node $i$ will be zero. 
\item The $\textsc{P-PRS}$ values for nodes reachable from $u$ will be equal to $\textsc{P-PRS}$ values calculated for a reduced network $G_u$, obtained from the original network by only accounting for the subset of nodes reachable from $u$ and connections between them (lines 6--15 in Algorithm \ref{algo:PPR}).
\end{itemize}

If the network is strongly connected, $G_u$ will be equal to the original network, yielding no change in performance compared to the original $\textsc{P-PRS}$ algorithm. However, if the resulting network $G_u$ is smaller, the calculation of $\textsc{P-PRS}$ values will be faster as they are calculated on $G_u$ instead of on the whole network. In our implementation, we first estimate if network $G_u$ contains less than $50\%$ (i.e. spread percentage) of nodes of the whole network (lines 6--14 in Algorithm \ref{algo:PPR}). This is achieved by expanding all possible paths from node $i$ and checking the number of visited nodes in each step. If the number of visited nodes stops increasing after a maximum of $15$ steps, we know we have found a network $G_u$, and we count its nodes. If the number of nodes is still increasing, we abort the calculation of $G_u$. We limit the maximum number of steps because each step of computing $G_u$ is computationally comparable to one step of the power iteration used in the PageRank algorithm \cite{pagerank1999} which converges in about $50$ steps. Therefore we can considerably reduce the computational load if we limit the number of steps in the search for $G_u$. Next, in lines 16--20, the \textsc{P-PRS} algorithm shrinks the personalized rank vectors based on non-zero values obtained as the result of the shrinking step.

\begin{figure}[htb!]
    \centering
    \includegraphics[width = \linewidth]{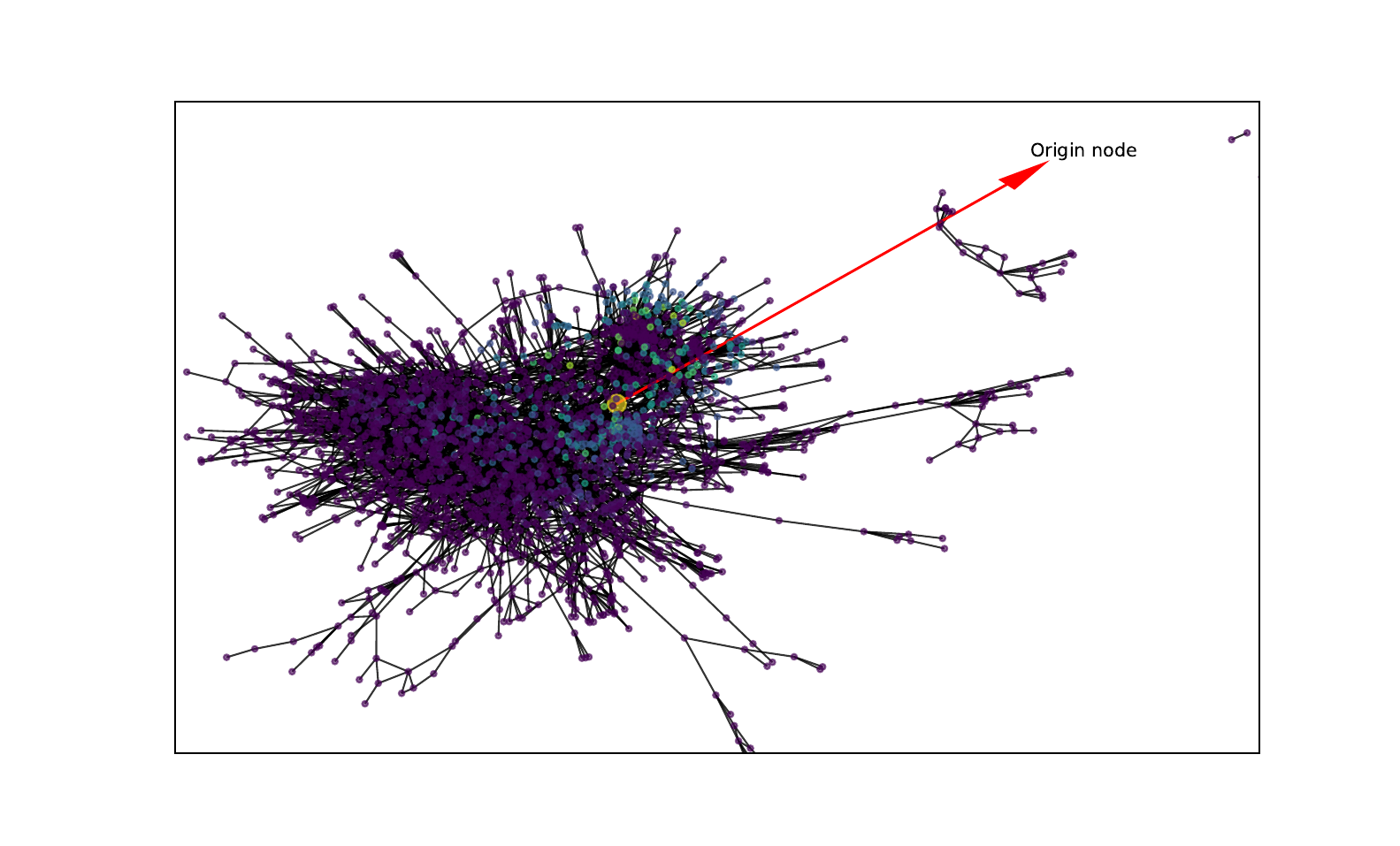}
    \caption{{Stationary distribution, visualized (\emph{Cora}). The origin node is the large yellow node pointed from. The upper right part of the network contains nodes that a simulated walker (from the origin node) likely ends up at (green-colored nodes).}}
    \label{fig:ppr-example}
\end{figure}

\noindent {\bf P-PRS computation step.}
In the second part of the algorithm (lines 23--38), node ranks are computed using the power iteration (Eq.~\ref{eqPR}), whose output consists of P-PRS vectors. An example stationary distribution is shown in Figure~\ref{fig:ppr-example} for the \emph{Cora} network.

For each node $u \in V$, a feature vector $\gamma_u$ (with components $\gamma_u(i)$, $1\leq i \leq |N|$) is computed by calculating the stationary distribution of a random walk, starting at node $u$. The stationary distribution is approximated using power iteration, where the $i$-th component $\gamma_{u}(i)^{(k)}$ of approximation $\gamma_u^{(k)}$ is computed in the $k+1$-st iteration as follows:
\begin{align}
\label{eqPR}
  \gamma_{u}(i)^{(k+1)} = \alpha \cdot \sum_{j \rightarrow i}\frac{\gamma_{u}(j)^{(k)}}{d_{j}^{out}}+(1-\alpha) \cdot v_{u}(i);k = 1, 2,\dots
\end{align}
\noindent The number of iterations $k$ is increased until the visit distribution converges to the final stationary distribution vector (P-PRS value for node $i$).
In the above equation, $\alpha$ is the damping factor that corresponds to the probability that a random walk follows a randomly chosen outgoing edge from the current node rather than restarting its walk. The summation index $j$ runs over all nodes of the network that have an outgoing connection {towards} $i$, (denoted as $j \to i$ in the sum), and $d_{j}^{out}$ is the out-degree of node $d_{j}$. Term $v_{u}(i)$ is the restart distribution that corresponds to a vector of probabilities for a walker's return to the starting node $u$, i.e. $v_{u}(u) = 1$ and $v_u(i)=0$ for $i\neq u$. This vector guarantees that the walker will jump back to the starting node $u$ in case of restart.\footnote{If the binary vector was composed exclusively of ones, the iteration would compute the global PageRank vector, and Eq.~\ref{eqPR} would reduce to the standard PageRank iteration.}

In a single iteration ($k \rightarrow k+1$), all stationary  distribution vector components  $\gamma_{u}(i)$, $1 \leq i \leq |N|$, are updated which result in the P-PRS vector $\gamma_{u}^{(k+1)}$. Increasing $k$ thus leads to the $\gamma_{u}^{(k)}$ eventually converging to the PageRank $\gamma_u$ of a random walk starting from node $u$ (see Algorithm~\ref{algo:PPR}). Eq.~\ref{eqPR} is optimized by using the \emph{power iteration}, which is especially suitable for large sparse matrices, since it does not rely on spatially expensive matrix factorization in order to obtain the eigenvalue estimates.\footnote{The power iteration (Eq.~\ref{eqPR}) converges exponentially, that is, the error is proportional to $\alpha^{k}$, where $\alpha$ is the damping factor and $k$ is the iteration number.}

The \textsc{P-PRS} algorithm simulates a first-order random walk in which no past information is incorporated in the final stationary distribution. The time complexity of the described \textsc{P-PRS} algorithm with shrinking for $k$ iterations is $\mathcal{O}(|N|(|E|+|N|)\cdot k)$ for the whole network, and $\mathcal{O}((|E|+|N|)\cdot k)$ for a single node.

\subsection{Additional shrinking by rank-based pivoting}
\label{sec:pivoting}

{We next present an additional step of shrinking explored as part of this work that offers scaling to very large networks. Recall (Algorithm~{\ref{algo:PPR}}) that the PageRank iteration, if the network is reduced, operates on the smaller adjacency matrix indexed via the set of nodes $\textrm{toReduce}$. This step, as offered in the Algorithm ~{\ref{algo:PPR}}, prunes out the nodes unreachable via traversal from the current node. This step preserves the computed vectors' properties, however, it does not guarantee asymptotically faster computation and is largely dependent on a given network's structure. For DNR to scale to very large networks, a more lossy selection scheme can be adopted. Recall the  $\textrm{toReduce}$, the set of nodes that define the final set of iterations that yield a given node's P-PRS-based representation. The idea discussed next defines the set $\textrm{toReduce}$ upfront; the size of this set is parametrized with an integer value $p$ (number of \emph{pivot nodes}). Members of this set are obtained as follows. We hypothesize that two main types of \emph{pivot nodes} need to be preserved in $\textrm{toReduce}$; namely, the nodes local to the node of interest, but also global nodes (via their relation to the node of interest). To address both concerns, we first define a given target node $u$'s neighbors as $\textrm{Ne}(u)$. Next, we define with $\textrm{argSortDes}(\textrm{PR(G)})$ the set of initial nodes, sorted by their PageRank values in descending order. Note that this step takes only $\mathcal{O}(|N|\log|N| + |E|)$ steps, and as such, scales to very large networks. The final set of $\textrm{toReduce}$ is constructed by first including all nodes from the neighborhood, followed by the global nodes which are not already in the neighborhood until the set is of cardinality $p$. We can formally define the set of \textrm{toReduce}$_u$ pivot nodes as the first $p$ nodes of the ordered union of the neighborhood and the top-ranked nodes, i.e.
}
\begin{align*}
    T^u &= \{a \in \textrm{Ne}(u),b \in  \textrm{argSortDes}(\textrm{PR(G)}) \setminus \textrm{Ne}(u) \} \\
    \textrm{toReduce}_u &= \{T_1^u, T_2^u, \dots, T_p^u\}.
\end{align*}
\noindent {Here, $T^u$ is the ordered union set (combined ordered sets), and $T_i^u$ the $i$-th element of that set (obtained with respect to node $u$). This heuristic selection of pivot nodes implies local neighborhood for low values of $p$, and mixed neighborhood for larger $p$ values.}
The computed (symbolic) node representations are used as inputs to the subsequent step of (neural) representation compression.

{The advantage of the deep neural network architecture discussed in the following section is that it can learn incrementally, from small batches of calculated }\textsc{P-PRS}{ vectors. In contrast, the previously developed HINMINE approach} \cite{kralj2017hinmine} {requires that all }\textsc{P-PRS}{ vectors are calculated prior to learning, which is due to HINMINE using the }$k${-nearest neighbors and support vector machine classifiers. This incurs substantial space requirements as the }\textsc{P-PRS} {vectors for the entire network require }$\mathcal{O}(|N|^{2})${ space. The DNR algorithm presented here uses a deep neural network instead, which can take as small input batches of }\textsc{P-PRS} {vectors. Therefore, only a small percentage of vectors need to be computed before the second step of the algorithm}(\emph{neural network training}) {can begin. This results in improved {space} and time complexities of the learning process.}
\subsection{{Node representation learning}}
\label{sec:architecture}

{In this section we address the second step of Deep Node Ranking algorithm (outlined in Figure~{\ref{scheme}}) -- the incremental compression of batches of personalized PageRank vectors via neural network learning. We next discuss the key formalisms used to describe the two types of learning implemented as part of DNR. We can formalize the key idea underlying DNR as the following mapping}

\begin{equation*}
    \textsc{DNR}: \underbrace{\mathbb{R}^{|N| \times |N|}}_{\textrm{Adjacency matrix}} \xrightarrow[]{\text{\textsc{P-PRS}}} \underbrace{[0,1]^{|N| \times |N|}}_{\textrm{\textsc{P-PRS} vectors}} \xrightarrow[]{\text{\textsc{NN}}_f}\underbrace{\mathbb{R}^{|N| \times d}}_{\textrm{Node embeddings}},
\end{equation*}
\noindent {where $d$ represents a \emph{latent dimension}, $N$ the set of nodes and $\textsc{DNR}$ the mapping approximated by the proposed approach. Note how the second space consists of visit \emph{probabilities} with respect to individual nodes. We will next focus on the two mapping methods displayed in the scheme above; $\textsc{P-PRS}$ and $\textsc{NN}_f$.}

{The first mapping ($\textsc{P-PRS}$) takes as input the network adjacency matrix and, if executed for each node, outputs the same dimensional matrix which contains richer, walk convergence-based information describing individual nodes (instead of only their neighbors). The initial adjacency can be stored as a sparse data structure, requiring only $\mathcal{O}(|E|)$ space. However, the probability matrix is commonly dense (with the exception of nodes in different components, for example), $\mathcal{O}(|N|^2)$ space can already pose a problem to the method's utility. To address this concern, we can consider \emph{batches} of nodes ($b$) from the first mapping onwards, potentially at no point requiring the full dense $\mathcal{O}(|N|^2)$ matrix. The first mapping adheres to this implementation due to the fact that with respect to individual nodes, \textsc{P-PRS} vectors can be obtained \emph{independently}. We denote with $\textsc{P-PRS}_b$ a produced \emph{batch} of such vectors. The union of all such batches (forming the set $B$) can be used to construct the whole probability matrix, i.e.}
\begin{equation}
    [0,1]^{|N| \times |N|} = \bigg \vert_{b \in B} \textsc{P-PRS}_b(A),
    \label{req:mapFirst}
\end{equation}
\noindent{ where $A$ represents the adjacency matrix and $\vert_b$ the concatenation alongside the first axis. Here, we assume the batches preserve the order of input nodes. The same property holds for transforming the $b \times |N|$-dimensional vectors into $b \times d$ dimensional ones with a \emph{trained} neural network $\textsc{NN}_f^b$. Similarly to the E.q.{~\ref{req:mapFirst}}, the final matrix can be written as}
\begin{equation*}
    \mathbb{R}^{|N| \times d} = \bigg \vert_{b \in B} \textsc{NN}_f^b(\textsc{P-PRS}_b(A)).
    \label{req:mapFirst2}
\end{equation*}

{The two equations above assume projection-ready mapping methods ($\textsc{NN}_f$ and \textsc{P-PRS}). The probability vector computation $\textsc{P-PRS}$ indeed requires no additional training. However, this is not the case for the neural network $\textsc{NN}$. Note that we denoted with $\textsc{NN}_f$ only the forward pass up to the hidden layer with $d$ outputs -- the embedding. To describe the whole process, the missing point remains the \emph{neural network training}. Denoted with $\textsc{NN}$, we represent a single epoch (forward and backward pass) of training the neural network (for all nodes). If we denote with $\omega$ the number of epochs required to train either an autoencoder-like, or an end-to-end architecture ($d$ is in this case the number of classes), DNR requires $\mathcal{O}(\omega \cdot (\textsc{NN} + |N| (|E| + |N|))$ operations. Furthermore, if the whole probability matrix fits into memory, the second product is decoupled, making the full probability matrix computed only once, resulting in complexity $\mathcal{O}(\omega \cdot \textsc{NN} + |N| (|E| + |N|))$. The initial complexity which re-computes the rank vectors for each batch ($b$) has, compared to the pre-computed rank matrix version lower space complexity, i.e. $\mathcal{O}(b \cdot |N|) << \mathcal{O}(|N|^2)$. This analysis demonstrates that for larger networks, additional computation needs to be performed in order to maintain the DNR's space complexity. Finally, the node pivoting scheme similarly reduces the space complexity of the rank matrix computation from $\mathcal{O}(|N|^2)$ to $\mathcal{O}(|N|\cdot p)$ ($p$ is the number of pivot nodes). Similarly, the time complexity reduces linearly ($|N| \rightarrow p$) for the ranking step. Hence, the pivoting scheme was hypothesized to improve both space and time-related performances substantially. Having discussed the coupled and the de-coupled (memoization) variants of DNR, it is apparent that the low space version will take much longer to compute compared to the memory-intensive version. To fine-tune this to a given hardware setting, DNR is able to estimate the approximate RAM utilization by assuming 32-bit floating point precision and takes as hyperparameter an integer number denoting the upper RAM bound. Should this bound be exceeded, the memory-efficient version is considered, and the faster one otherwise. In this work, we set this bound to 16GB.}

\subsection{{DNRNet: A neural network architecture}}
\label{sec:DNRNet}
{In the previous section, a description of the core feature construction process based on personalized node ranking was described alongside its time and space complexities. We next discuss in more detail the considered neural network architecture and the training regime, which is also a contribution of this work.}

\begin{figure}[htb!]
    \centering
    \includegraphics[width = .6\linewidth]{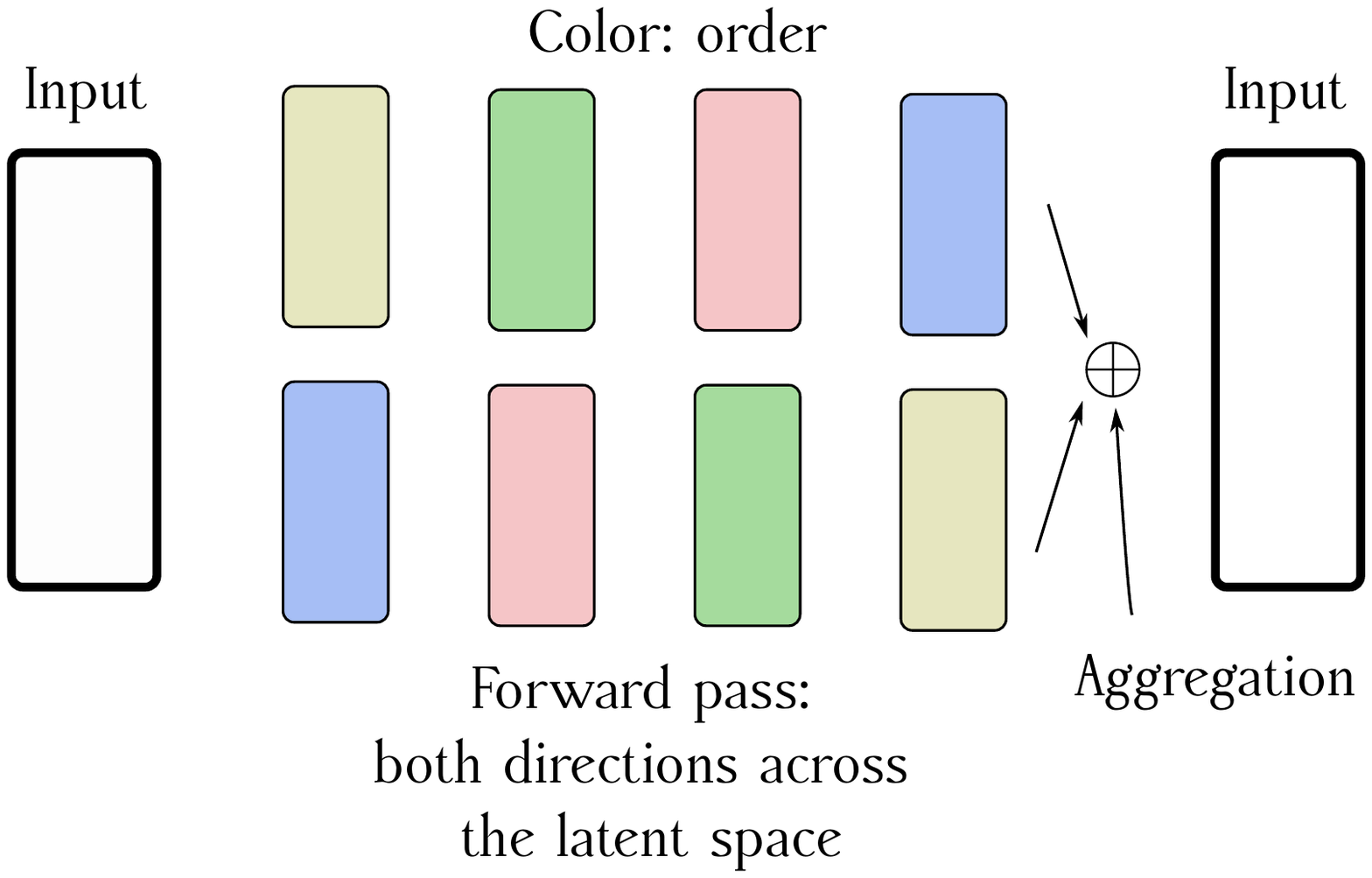}
    \caption{{The forward pass with the inclusion of reversed latent spaces.}}
    \label{fig:scheme-fp}
\end{figure}

{We are interested in compressing the \textsc{P-PRS}-based representation (Equation~{\ref{req:mapFirst}}) we henceforth refer to as $\boldsymbol{P}$. The goal of the designed neural network is to compress this representation from dimension $|N|$ to $d$, in \emph{unsupervised} manner. To achieve this compression, we implemented an autoencoder-like architecture with a forward pass defined as follows (note the indexing).}
\begin{align*}
    l_i &= \textrm{Dropout}(\textrm{ELU}(\boldsymbol{W_d}^T \cdot \boldsymbol{P} + b_i))\\
    h_1 & = \textrm{ELU}(\boldsymbol{W_{d1}}^T \cdot l_i + b_{h1}) \\
    &\dots  && \text{Initial embedding order}\\
    h_k & = \textrm{ELU}(\boldsymbol{W_{dk}}^T \cdot h_{k-1} + b_{hk})\\
    r_1 & = \textrm{ELU}(\boldsymbol{W_{dk}}^T \cdot l_i + b_{r1})\\
    &\dots && \text{Reversed embedding order} \\
    r_k & = \textrm{ELU}(\boldsymbol{W_{d1}}^T \cdot r_{k-1} + b_{rk})\\
    l_o &= \boldsymbol{W_o}^T (\frac{1}{2} \cdot r_k \oplus h_k).\\
\end{align*}
\noindent {The first part of the architecture projects and activates the input probabilities ($\boldsymbol{P}$) to a lower dimension ($d$). 
The ELU activation is defined as }
\begin{equation*}
    \textrm{ELU}(x) =
        \begin{cases}
        x; x > 0 \\
        \alpha \cdot (e^{x} - 1);  x \leq 0
        \end{cases}.
\end{equation*}
{The parameter $\alpha$ was set to 1 throughout this work.
The inner part of the architecture consists of multiple same-dimensional ($d$) layers, which refine the representation. The final layer projects the refined representation back to the initial dimension ($|N|$). The key component is the regularization (dropout) prior to the embedding layers, as it notably improved the architecture's stability during design. The loss function used is the Smooth L1 Loss defined as:}
\begin{equation*}
    \mathcal{L}_n = \begin{cases}
    \frac{1}{2} \cdot (x_n - y_n)^2/\beta; |x_n - y_n| < \beta \\
    |x_n - y_n| - \frac{1}{2} \cdot \beta; \textrm{otherwise}
    \end{cases}
\end{equation*}
\noindent
{
Here, $x_n$ represents the prediction, $y_n$ the actual value and $\beta$ a parameter (set to 1.0 in this work). The loss is averaged on the batch level.
The key novelty of the proposed neural network-based compression is not the architecture but the way \emph{forward passes are conducted}. We impose an additional constraint on the intermediary representations by implementing the forward pass so that it includes multiplication \emph{in both ways} across the hidden layers (note the shared parameters -- there is no weight duplication). This means that each forward pass incorporates a scenario where the first hidden layer is first, but also last in the forward pass (inverted latent space); with this, we enforce reverse consistency, as \emph{all} intermediary representations are used to obtain the final embedding. This is possible due to the symmetric nature of the activation-dense layers -- inverting the order during the forward pass amplifies the effect of different hidden layers with the same type of output. The $\oplus$ denotes the Hadamard summation (elementwise). Note that such inverse projections during the same forward pass are possible because the dimensionalities of the intermediary representations are all the same ($d$). A schematic overview of this idea is shown in} Figure~\ref{fig:scheme-fp}.

\begin{figure}[htb!]
    \centering
    \includegraphics[width = .6\linewidth]{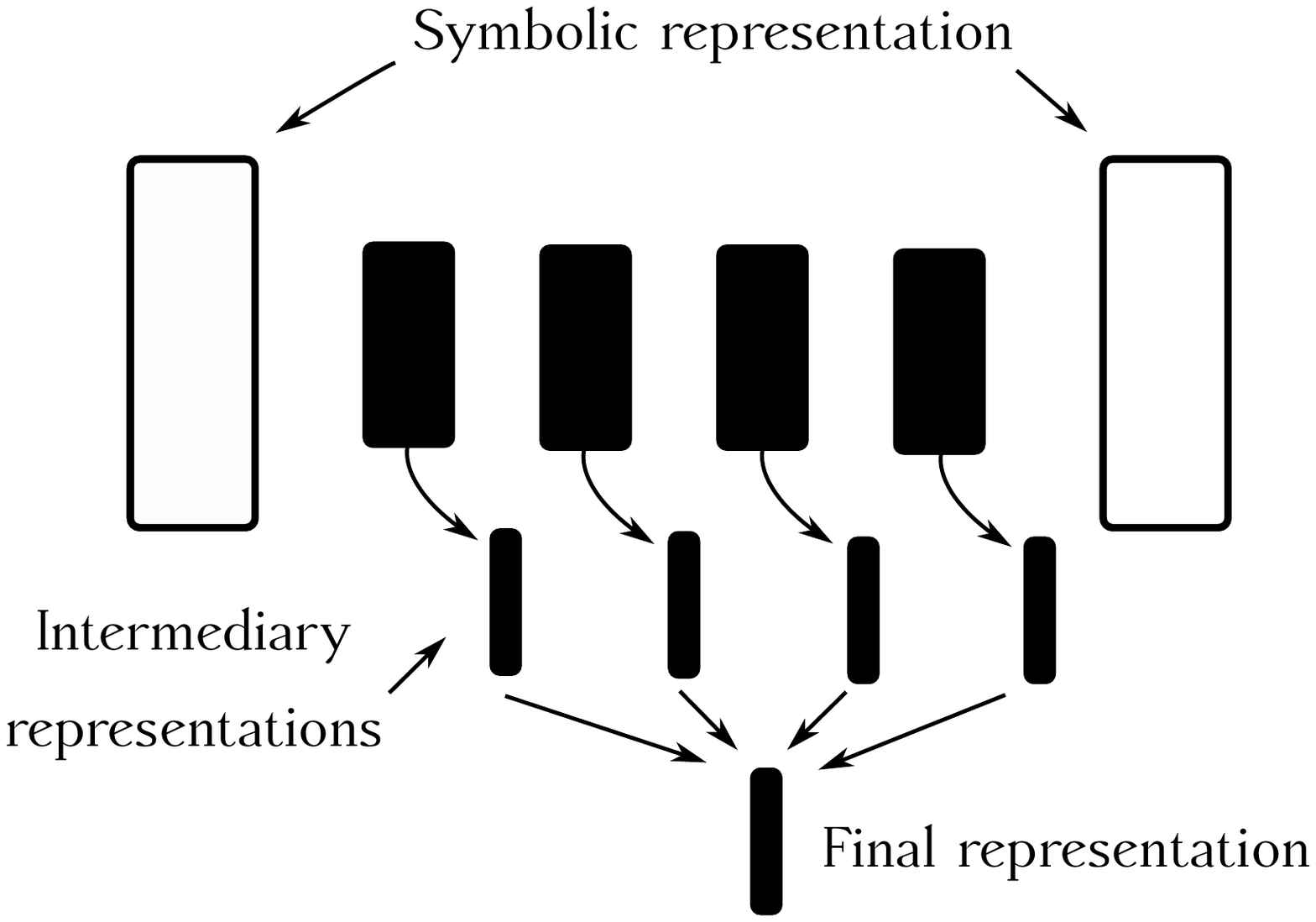}
    \caption{{DNRNet's representation construction process. The final representation ($\boldsymbol{E}$) is obtained as an aggregate of \emph{all} relevant intermediary layers.}}
    \label{fig:scheme-representation}
\end{figure}

{Lastly, we discuss how the final representations (node embeddings) are obtained. Recall that $h_1, \dots, h_k$ represent the outputs of the intermediary embedding layers. The final node representations ($\boldsymbol{E}$) are obtained by performing Hadamard summation across these intermediary representations and dividing with the number of hidden outputs, i.e.,}
\begin{equation*}
    \boldsymbol{E} = k^{-1} \cdot \bigoplus_i \boldsymbol{h}_i.
\end{equation*}
{Schematic overview of this process is shown in Figure}~\ref{fig:scheme-representation}. {The main reason such multi-space aggregation is conducted is that the information used for reconstructing the origin rank space is likely distributed across all hidden layers, implying that by considering, e.g., only the last layer, valuable parts of the final representation could be lost. This idea was inspired by how representations are obtained from contextual language models{~\cite{reimers-2019-sentence-bert}}.}

\section{Data sets and experimental setting}
\label{sec:experimental}

This section first describes the data sets used, the experimental setting and the DNR implementations tested together with their hyperparameters, followed by a description of the compared baseline approaches.

\subsection{Data sets}
We evaluated the proposed approach on \finaldatanum real-world complex networks, three of them introduced in this work, which is one of the largest collections
of complex networks {for the task of} node classification. The \emph{Homo Sapiens} (proteome) \cite{stark2010biogrid}, POS tags \cite{mahoney2011large} and Blogspot data sets \cite{zafarani2009social} are used in the same form as in \cite{grover2016node2vec}.

The \emph{Homo sapiens} data set represents a subset of the human proteome, i.e. a set of interacting proteins. The sub-network consists of all proteins for which biological states are known \citep{stark2006biogrid}. The goal is to predict protein function annotations. 
The POS data set represents part-of-speech tags obtained from Wikipedia---a co-occurrence network of words appearing in the first million bytes of the Wikipedia dump \citep{mahoney2011large}. Thus, different POS tags are predicted.
The Blogspot data set represents a social network of bloggers (Blogspot website) \citep{zafarani2009social}. The labels
represent bloggers' interests inferred through the metadata
provided by the bloggers.
 The CiteSeer citation network consists of scientific publications classified into one of the six classes (categories) \citep{lu2003link}.
The Cora citation network consists of scientific publications classified into one of seven classes (categories) \citep{lu2003link}.
The E-commerce network is a heterogeneous network connecting buyers with different products. As DNR and the compared baseline algorithms operate on homogeneous networks, the E-commerce network was transformed to a homogeneous network prior to learning using a term frequency weighting scheme~\cite{kralj2017hinmine}. The created edges represent mutual purchases of two persons, i.e. two customers are connected if they purchased an item from the same item category.
We refer the interested reader to~\cite{kralj2017hinmine} for a detailed description of the data set and its transformation to a homogeneous network. The two-class values being predicted correspond to the buyers' gender. {The \emph{film}, \emph{squirrel},\emph{chameleon}, \emph{wisconsin}, \emph{texas} and \emph{cornell} data sets are based on a recent study about geometric deep learning{~\cite{pei2020geom}}. Given that some of the data sets have features, and the purpose of this paper is structure-only learning, instead of neglecting the feature spaces, we converted them into weights between nodes as follows. If the cardinality of the feature spaces of a given node was the same for all nodes, we computed the weights as inverse Euclidean distances with one added to the denominator (similarities). If the features were sets, we computed the weights as cardinalities of the intersection sets between pairs of nodes.}

One of the contributions of this work is also three novel node classification data sets, which we constructed as follows.
Two data sets are related to Bitcoin trades \cite{kumar2016edge}. The two networks correspond to transactions within two different platforms, namely Bitcoin OTC and Bitcoin Alpha. Each edge in this network represents a transaction along with an integer score denoting trust in the range $[-10,10]$ (zero-valued entries are not possible). We reformulate this as a classification problem by collecting the trust values associated with individual nodes and considering them as target classes. The resulting integer values can thus belong to one of the 20 possible classes. Note that more than a single class is possible for an individual node, as we did not attempt to aggregate trust scores for individual nodes.

The \emph{ions} data set is based on the recently introduced protein-ion binding site similarity network \cite{vskrlj2018insights}. The network was constructed by structural alignment using the ProBiS family of algorithms \cite{konc2014probis,konc2012parallel,konc2017genprobis} where all known protein-ion binding sites were considered. The obtained network was pruned for structural redundancy as described in \cite{vskrlj2018insights}. Each node corresponds to one of 12 possible ions, and each weighted connection corresponds to the ion-binding site similarity between the two considered {binding sites}. Overall, this is to date one of the largest collections of structure-only node classification benchmark data sets.

\begin{small}
  \begin{table}[t!]
    \centering
    \caption{{Networks used in this study and their basic statistics.}}
    \resizebox{0.95\textwidth}{!}{
\begin{tabular}{lccccccc}
\toprule
         Name & \#Classes & \#Nodes & \#Edges &  Mean deg &   CC &  CCoef &  Density \\
\midrule
      \emph{cornell} &        4 &    183 &    280 &      3.06 &    1 &   0.17 &   0.0168 \\
        \emph{texas} &        4 &    183 &    295 &      3.22 &    1 &   0.20 &   0.0177 \\
    \emph{wisconsin} &        4 &    251 &    466 &      3.71 &    1 &   0.21 &   0.0149 \\
         \emph{ions} &       12 &   1969 &  16092 &     16.35 &  326 &   0.53 &   0.0083 \\
    \emph{chameleon} &        4 &   2277 &  31421 &     27.60 &    1 &   0.48 &   0.0121 \\
         \emph{cora} &        7 &   2708 &   5278 &      3.90 &   78 &   0.24 &   0.0014 \\
     \emph{citeseer} &        6 &   3327 &   4676 &      2.81 &  438 &   0.14 &   0.0008 \\
\emph{Bitcoin\_alpha} &       20 &   3783 &  14124 &      7.47 &    5 &   0.18 &   0.0020 \\
 \emph{Homo\_sapiens} &       50 &   3890 &  38739 &     19.92 &   35 &   0.15 &   0.0051 \\
          \emph{POS} &       40 &   4777 &  92517 &     38.73 &    1 &   0.54 &   0.0081 \\
     \emph{squirrel} &        4 &   5201 & 198493 &     76.33 &    1 &   0.42 &   0.0147 \\
      \emph{Bitcoin} &       20 &   5881 &  21492 &      7.31 &    4 &   0.18 &   0.0012 \\
         \emph{film} &        4 &   7600 &  14056 &      3.70 & 1975 &   0.04 &   0.0005 \\
     \emph{Blogspot} &       39 &  10312 & 333983 &     64.78 &    1 &   0.46 &   0.0063 \\
 \emph{ecommerce\_tf} &        2 &  29999 & 178608 &     11.91 & 8304 &   0.48 &   0.0004 \\
\bottomrule
\end{tabular}
}
\label{tbl:stats}
\end{table}
\end{small}

The {considered} data sets are summarized in Table \ref{tbl:stats}. In the table, CC denotes the number of connected components. The clustering coefficient measures how nodes in a graph tend to cluster together and is computed as the ratio between the number of closed triplets and the number of all triplets. The network density is computed as the number of actual connections divided by all possible connections. The mean degree corresponds to the average number of connections of a node. Links to data sets, along with other material presented in this paper, are discussed in Section~\ref{availability}.

Furthermore, to test the DNR's scalability, we created \finalsynthetic Erd\H{o}s-R\'enyi networks in node range from 2{,}500 to 35{,}000 in the increments of 1{,}000 with different seeds and the probability parameter set to 0.05 (sparser networks).

\subsection{Experimental setting}
\label{sec-setting}

In this section, we describe the experimental setting used to evaluate the proposed method against the existing baselines.

There are two main evaluation aspects relevant to this paper; investigating the quantitative performance of embeddings on a given downstream task and computation time. To assess the classification performance, we use the same evaluation scheme as in related work on node classification~\citep{tang2015line,grover2016node2vec,perozzi2014deepwalk,netMF}. Here, as all methods for node embedding construction are unsupervised, an embedding is first constructed and used as input to a logistic regression-based classification scheme suitable for multiclass and multilabel classification tasks.

We repeated the classification experiments five times and averaged the results to obtain stable performance estimates with corresponding variabilities.
The performance of trained classifiers was evaluated {by} using micro and macro $F_1$ scores, as these two measures are used in the majority of related node classification studies \citep{tang2015line,grover2016node2vec,perozzi2014deepwalk,netMF}.

Due to many classifier comparisons, we utilize the Friedman test with Nemenyi \emph{post hoc} correction to compute the statistical significance of the differences. The results are visualized as critical difference diagrams, where {average} ranks of individual algorithms according to scores across all data set splits are presented~\cite{demsar2006}. The selected algorithms are also compared via Bayesian hierarchical t-test~\cite{bayesiantests2016} with a prior value of $\rho = 0.8$ and rope region value set to 2\%. All experiment repetitions were used for posterior sampling.

All experiments were conducted on a machine with 64GB RAM, 6 core Intel(R) Core(TM) i7-6800K CPU @ 3.40GH with a Nvidia 1080 GTX GPU. As the maximum amount of RAM available for all approaches was 64GB, the run is marked as unsuccessful, should this amount be exceeded. Further, we gave each algorithm at most five hours for learning the embeddings and subsequent classification. We selected these constraints as the networks used are of medium size, and if a given method cannot work on these networks, it will not scale to larger networks; e.g., social networks with millions of nodes and tens of millions, or even billions of edges without substantial adaptation of the method. The unsuccessful runs were replaced with a random embedding.
Node ranking was implemented {by} using sparse matrices from the SciPy module \cite{Virtanen2020} and the PyTorch library~\cite{NEURIPS2019_9015}.

\subsection{DNR implementations}
\label{sec:implementations}
{
We implemented the following variants of DNR, each emphasizing a different aspect of the algorithm.

The \textbf{DNR} represents a default DNR implementation with no node pivoting, two hidden layers, trained for at most 100 epochs with the stopping criterion of five epochs. The learning rate was set to 0.01 and adaptively decreased throughout the training. The upper memory bound was set to 16GB, meaning that networks that would require more space would be computed incrementally, on the fly, reducing the space but increasing the computation time. The latent dimension was for this and all other embedding-based methods set to 128 (as also seen in related work). The \textbf{DNR4} architecture includes four hidden layers instead of two, and \textbf{DNR8} eight hidden layers. The \textbf{DNRPH} is a DNR variant with the pivoting node number set to $|N|/2$. The \textbf{DNRPQ} to $|N| \cdot 0.75$ and \textbf{DNRPM} to $\sqrt{|N|}$. All pivot number estimates were rounded to the nearest integer. Finally, we implemented the symbolic-only learner we refer to as \textbf{DNR-symbolic}, which outputs the $\boldsymbol{E}_s \in |N|^2$ matrix of personalized rank vectors (symbolic part of full DNR).}

The P-PRS algorithm parameters (constant throughout all experiments) were set as follows. $\epsilon$, the error bound, which specifies the end of an iteration, was set to $10^{-6}$. Max steps, the number of maximum steps allowed during one iteration was set to 100{,}000 steps. Damping factor; the probability that a random walker continues at a given step was set to $0.5$. Spread step, the number of iteration steps allowed for the shrinking part was set to $10$. Spread percent, the maximum percentage of the network to be explored during shrinking was set to $50\%$.

\subsection{The baseline approaches}

We tested the proposed approach against different baselines outlined below.
The baselines were selected as they are currently considered as either very weak (random) or strong (node2vec, struc2vec). All approaches apart from label propagation are node embedding algorithms. For label propagation, the same data splits were used for classification evaluation as for the logistic regression when considering embedding-based learning.
\begin{itemize}
\item \textbf{node2vec} \cite{grover2016node2vec}. This algorithm maximizes the likelihood of preserving network neighborhoods of nodes. This is achieved via biased random walk sampling. This algorithm is considered a strong baseline for structure-only learning.
\item \textbf{struc2vec} ~\cite{struc2vec}. {This algorithm uses a hierarchy-like structure to measure node similarity at different scales and constructs a multilayer graph to encode structural similarities and generate structural context for nodes. It remains one of the key approaches capable of including information on structural similarity.}
\item \textbf{Label Propagation (LP)} \cite{zhu2002learning}. Label propagation is a well-known algorithm for node classification. It operates by incrementally sending information from the neighboring nodes to the unlabeled nodes, eventually reaching an equilibrium and yielding the final set of predictions for the masked part of the network.
\item \textbf{GraphWave}~\cite{donnat2018} {is a method that represents each node's local network neighborhood via a low-dimensional embedding by leveraging spectral graph wavelet diffusion patterns. This is one of the more scalable methods considered in this work.}\footnote{Python 3 implementation used: \url{https://github.com/benedekrozemberczki/GraphWaveMachine}}
\item \textbf{Graph Neural Networks} (GAT and GCN) \cite{velivckovic2018graph}. We trained the models with the stopping criterion of 100 epochs {for} up to 1000 epochs. Due to unstable performance, we report the best performance (epoch scoring best). Further, as GATs were not initially implemented for multilabel classification, we extended them, so they minimize binary cross-entropy and output a sigmoid-activated space with the same {dimension} as the number of targets (the multiclass version does not work for such problems). As this branch of models operates with additional features assigned to nodes, and the considered benchmark data sets do not possess such features, we used the identity matrix of the adjacency matrix as the feature space, thus expecting sub-optimal performance. This Algorithm was shown to outperform other variants of graph neural networks such as the GCNs \cite{kipf2016semi} which were also considered under the same training regime.
\item {\textbf{Random baseline} which is a random float matrix $\in [0, 1]^{|N| \times d}$. The PyTorch-Geometric library was used for the two baselines{~\cite{Fey/Lenssen/2019}}}

\end{itemize}
For all baselines, suggested default hyperparameter settings were used (either taken from papers or from the codebases). Similarly, default configurations of DNR variants were used to ensure fair comparisons (no additional hyperparameter optimization was conducted across data sets). Thus, we evaluated out-of-the-box performance -- additional hyperparameter tunning could significantly increase the training time and render some of the methods inapplicable even at the mid-scale networks considered in this work.

\section{Results}
\label{sec:results}
In this section, we present the empirical results and discuss their qualitative as well as quantitative aspects. We first present the results for the node classification task, followed by a qualitative evaluation of the proposed DNR algorithm for the task of node visualization.

\subsection{Classification performance}
\label{sec:cperf}
\begin{figure}[htb!]
    \centering
     \begin{subfigure}[b]{\textwidth}
         \centering
        \includegraphics[width = \linewidth]{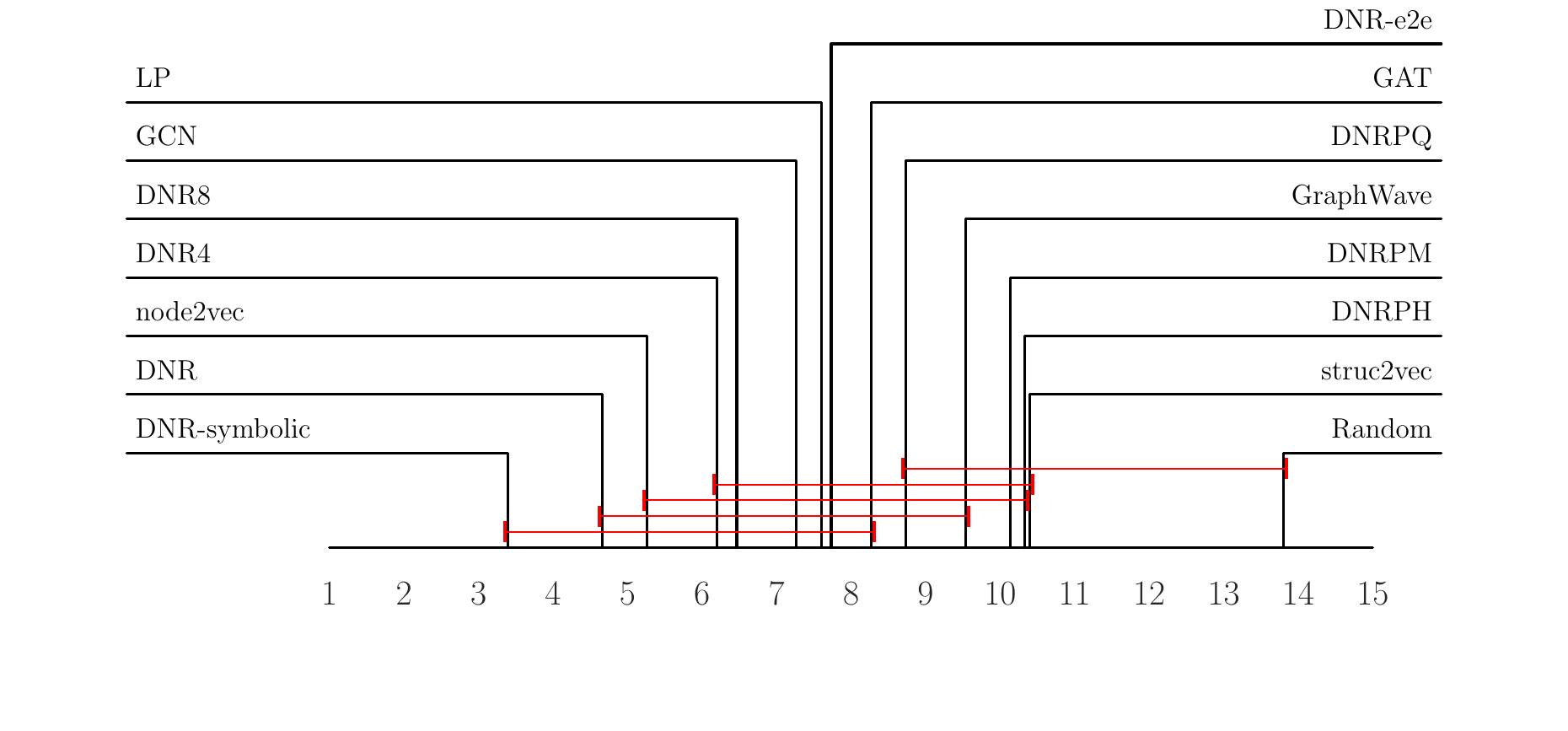}
        \caption{Critical differences -- micro F1.}
     \end{subfigure}
     \begin{subfigure}[b]{\textwidth}
         \centering
        \includegraphics[width = \linewidth]{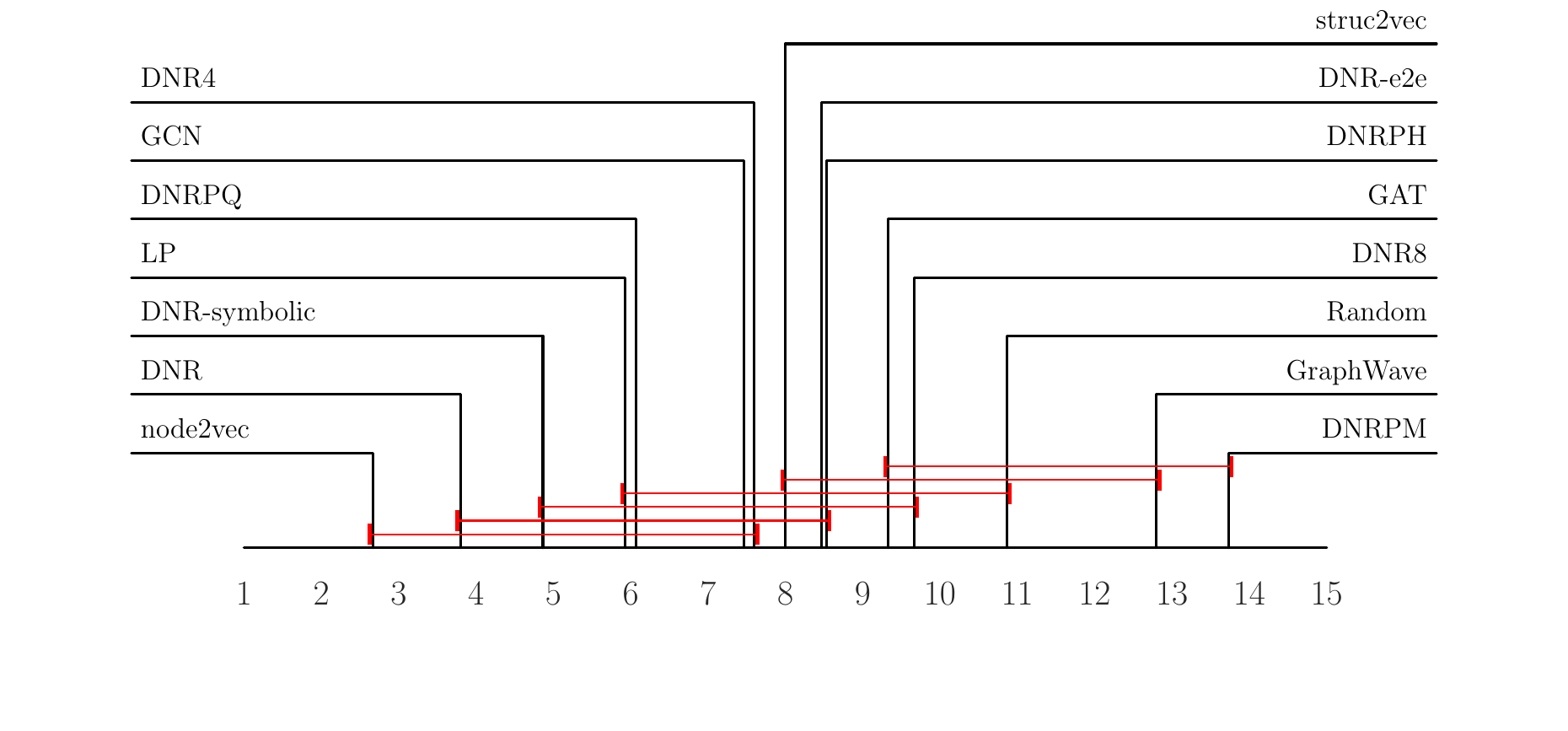}
        \caption{Critical differences -- macro F1.}
     \end{subfigure}
     \caption{Overview of classification performance -- critical difference diagrams.}
     \label{fig:overview-cd}
\end{figure}

\begin{figure}[htb!]
    \centering
    \begin{tabular}{cc}
\subcaptionbox{DNR and node2vec.}{\includegraphics[width = .5\linewidth]{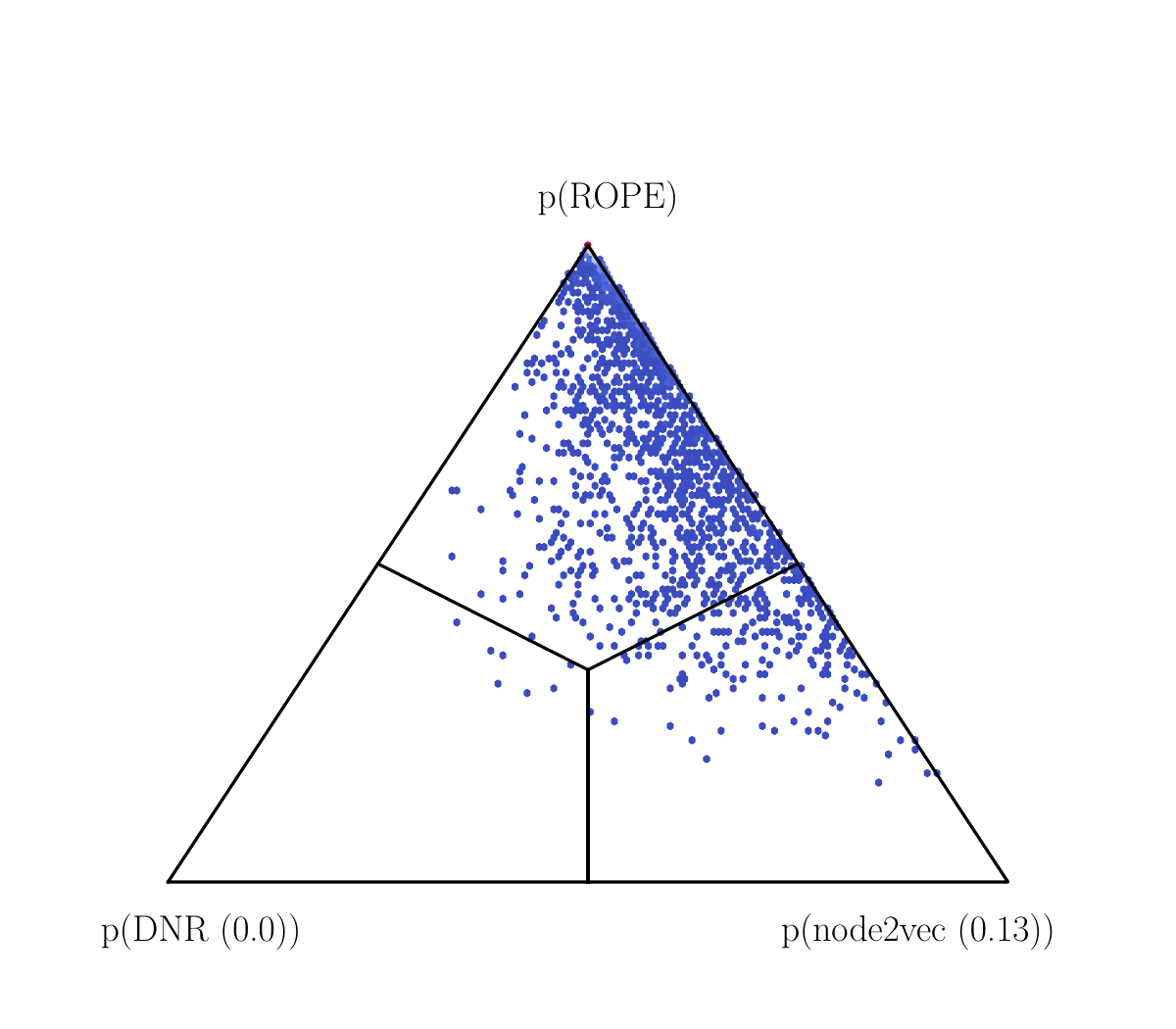}} &
\subcaptionbox{DNR and struc2vec.}{\includegraphics[width = .5\linewidth]{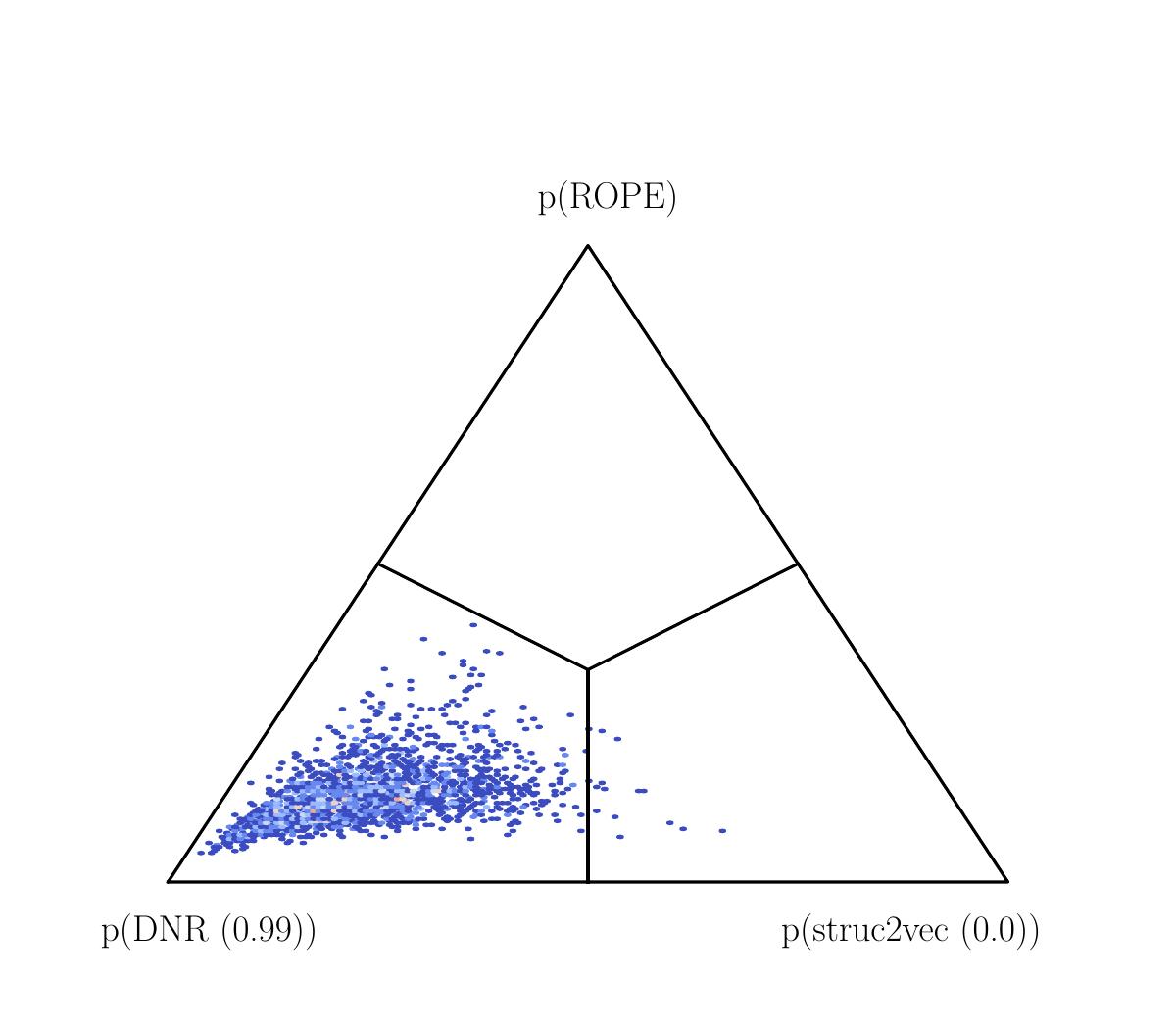}} \\
\end{tabular}
     \caption{Bayesian comparison of selected algorithm pairs.}
    \label{fig:bayesian}
\end{figure}

We first present the results of classification experiments. In Figure~\ref{fig:overview-cd} the reader can observe the critical difference plots of micro and macro F1 scores aggregated across all data sets.
It can be observed that similar algorithms dominate with respect to both scores; node2vec, DNR, DNR-symbolic and label propagation are amongst the best-performing ones. The differences between the best performers are insignificant, as demonstrated via statistical analysis (CD diagrams)~\cite{demsar2006}.
\begin{figure}[htb!]
    \centering
     \begin{subfigure}[b]{0.47\textwidth}
         \centering
        \includegraphics[width = \linewidth]{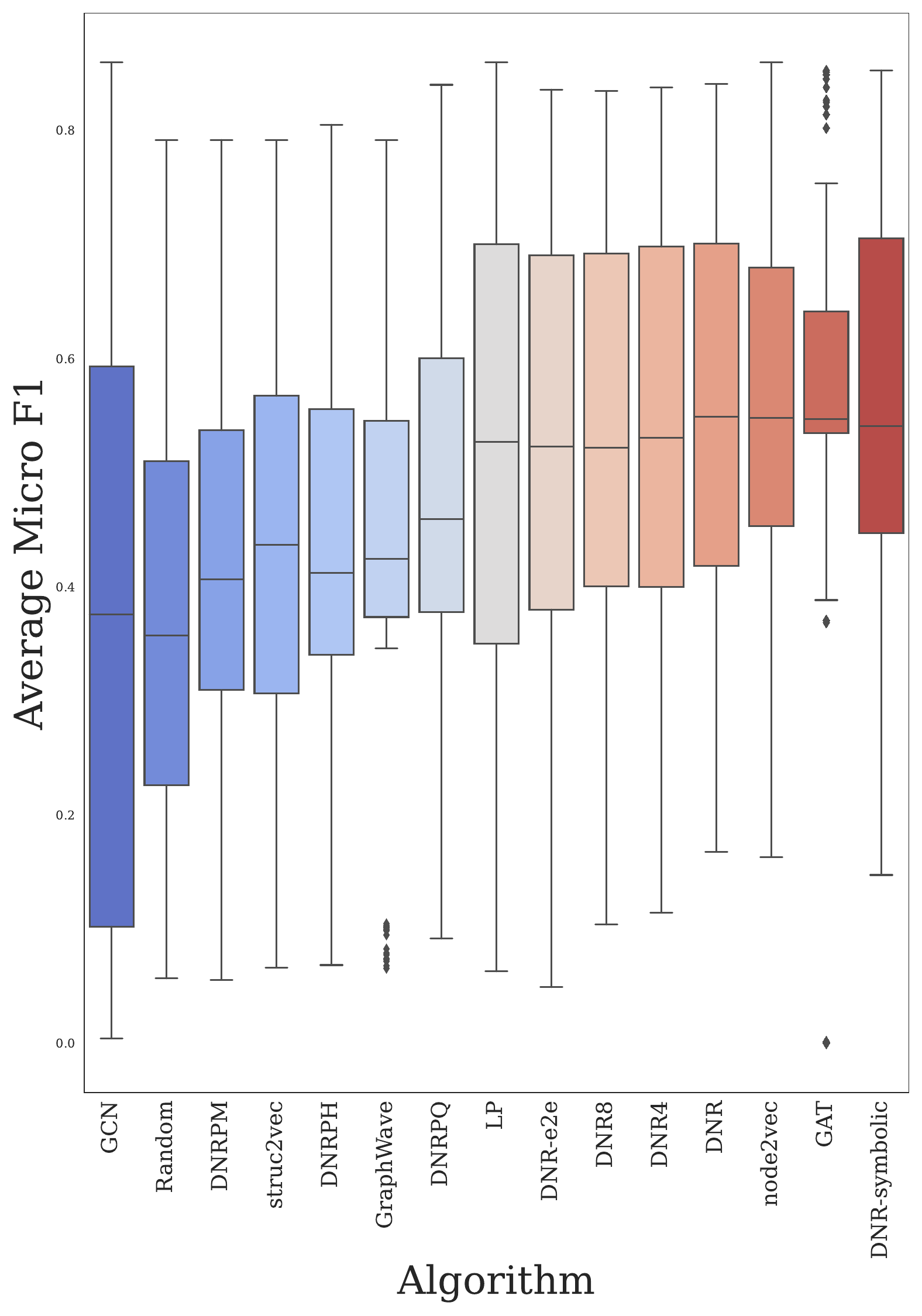}
        \caption{Box plots -- micro F1.}
     \end{subfigure}
     \begin{subfigure}[b]{0.47\textwidth}
         \centering
        \includegraphics[width = \linewidth]{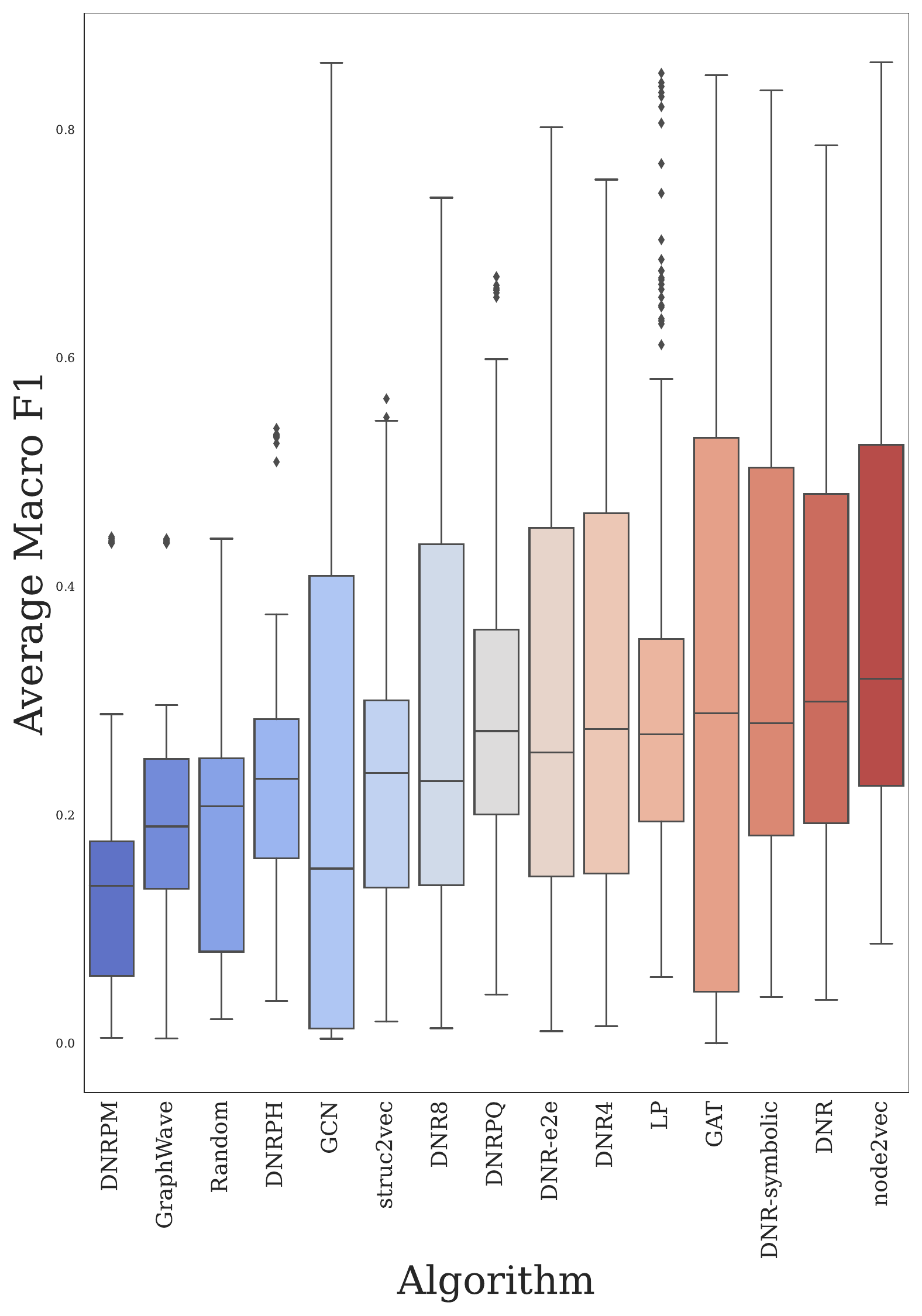}
        \caption{Box plots -- macro F1.}
     \end{subfigure}
     \caption{Micro and macro F1 performance distributions for considered algorithms.}
    \label{fig:overview-box}
\end{figure}
{Next, GraphWave and DNRPM underperform w.r.t. macro F1 (Figure{~\ref{fig:overview-box}}). This observation could be due to multiple factors, ranging from GraphWave's hyperparameter sensitivity, poor performance on small networks (too much information is lost) or similar. Amongst the best performing algorithms are either the default DNR variant with two hidden layers, DNR-symbolic or node2vec. The DNR variant implementing a deeper neural network (DNR8) performed worse than the more shallow versions, indicating overfitting (highly likely especially for smaller networks). The DNRPM variant, which uses a substantially reduced version of the adjacency matrix for rank computation, performed better than random when considering micro F1. However, it was overall amongst the worst performing variants of DNR. The DNRPQ variant performed better, indicating that node pruning can have a substantial impact on the final representation -- too low values of {$p$} indicate detrimental effects on the final performance. The end-to-end variant of DNR performed competitively w.r.t. micro F1; however, it performed worse when considering macro F1. This result indicates over-fitting, but also the method's potential sensitivity to the classification of nodes in smaller networks (see the appendix materials for detailed scores on smaller networks). Note that the proposed end-to-end DNR out-performed the two GNN baselines. Current results indicate that structure-only learning is harder for GCN and GAT-based models -- either due to higher possibility of overfitting or due to space complexity which arises if considering the the attention-based architecture.
The results indicate that the neural network in neuro-symbolic DNR variants, as expected, acts as a compression layer, losing some of the expressive power of the origin rank space at the cost of being more efficient space-wise.}
Bayesian comparison of default DNR with node2vec and struc2vec confirms the results obtained via frequentist analysis and is shown in Figure~\ref{fig:bayesian}.
The numbers denote the posterior probability estimates (higher is better). Note the insignificant difference between DNR and node2vec (most of the density is in the rope region), but the significant difference (as also confirmed via frequentist analysis) between DNR and struc2vec. Overall, the Bayesian analysis confirms the findings supported by the classical analysis.

\subsection{Execution time analysis}
\begin{figure}[htb!]
    \centering
    \begin{tabular}{cc}
\subcaptionbox{Execution time.}{\includegraphics[width = .5\linewidth]{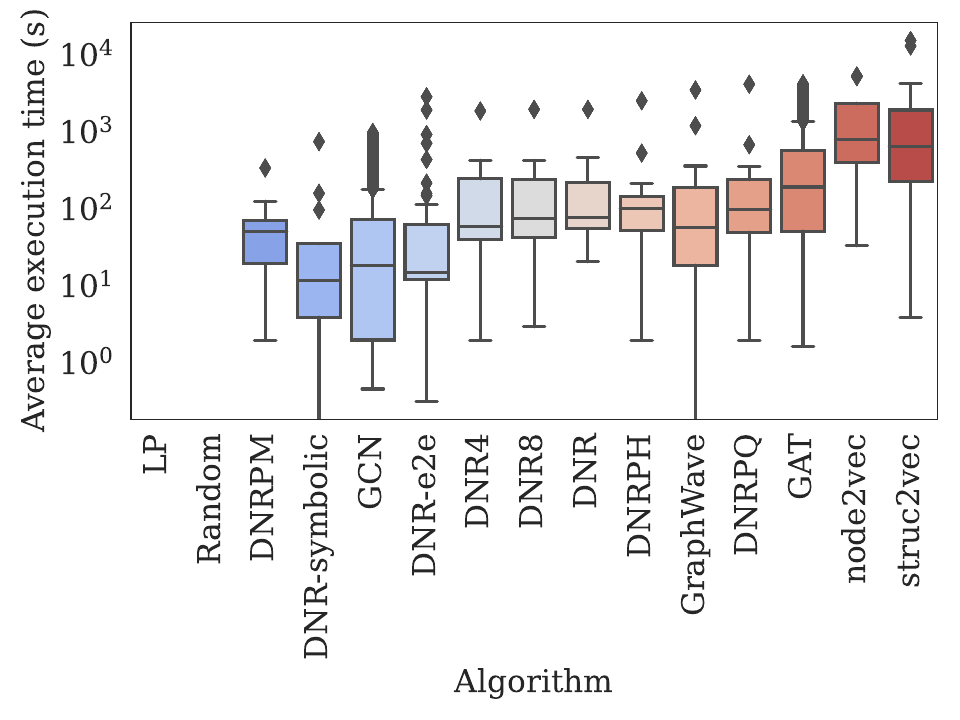}} &
\subcaptionbox{Time per data set.}{\includegraphics[width = .5\linewidth]{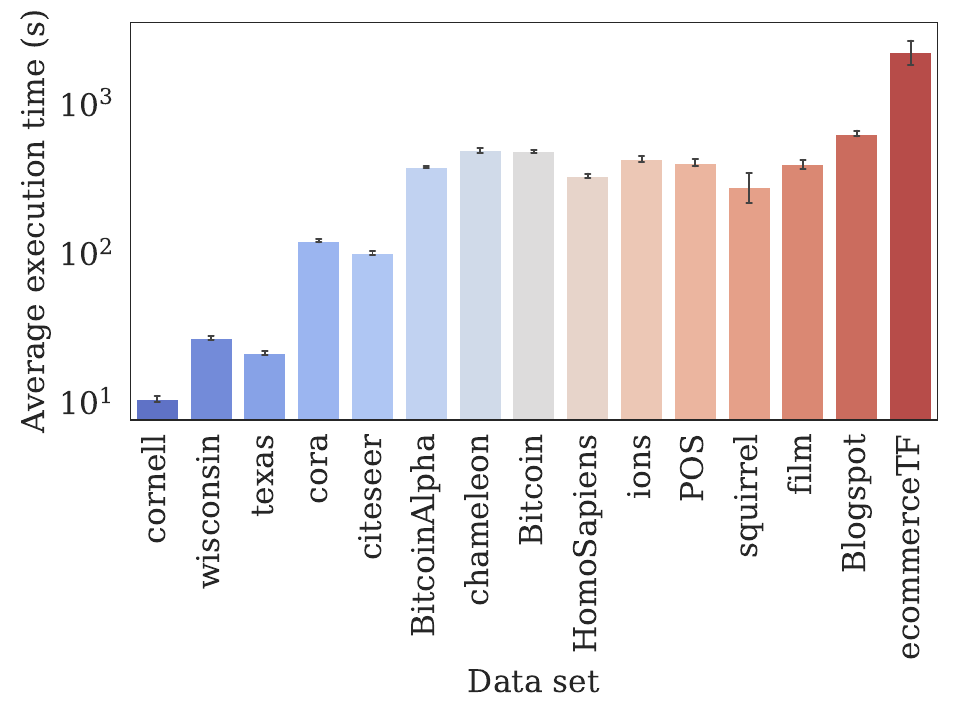}} \\
\end{tabular}
     \caption{Execution time analysis. The proposed DNR algorithm performs substantially faster than struc2vec and node2vec.}
    \label{fig:execution-time1}
\end{figure}

Overview of the execution times is shown in Figure~\ref{fig:execution-time1}. We present overall execution times followed by the per-dataset execution times.

It can be observed that the fastest DNR variants perform up to two orders of magnitude faster than, e.g., struc2vec and node2vec. Given that, e.g., DNR offers very similar performance, this result serves as a strong case for using DNR-based embeddings, especially on larger networks. Further, note that the execution time includes both embedding construction and classification, rendering the random baseline slower than label propagation (logistic regression is the bottleneck in this case).

\subsection{Number of pivot nodes and scalability}
\begin{figure}[htb!]
    \centering
     \begin{subfigure}[b]{0.47\textwidth}
         \centering
        \includegraphics[width = \linewidth]{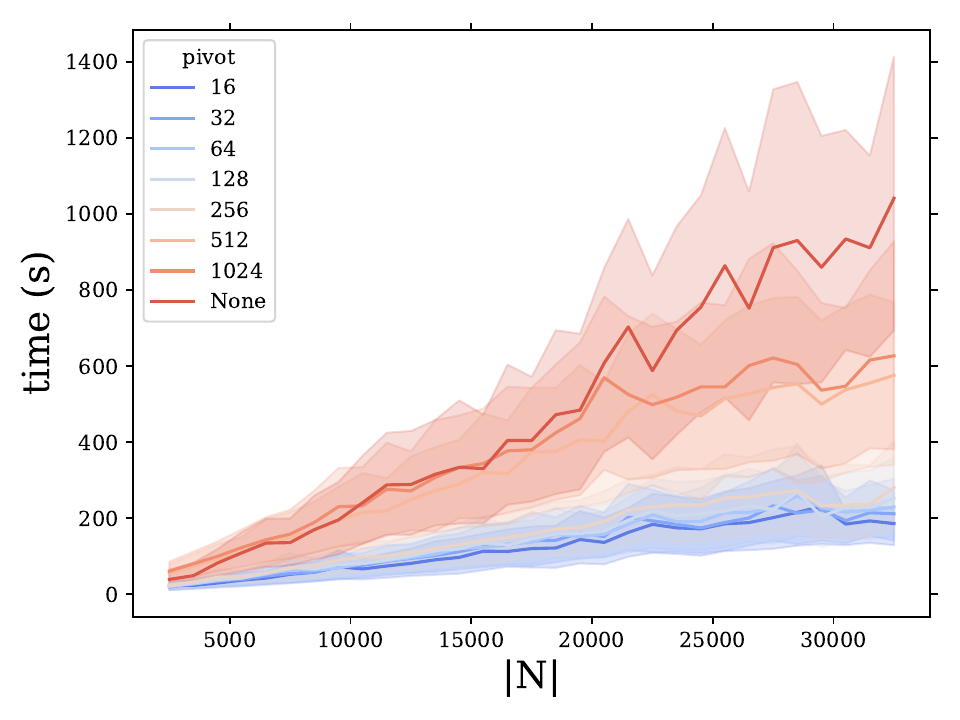}
        \caption{Time w.r.t. $|N|$.}
         \label{fig:scalability-abl}
     \end{subfigure}
     \begin{subfigure}[b]{0.47\textwidth}
         \centering
        \includegraphics[width = \linewidth]{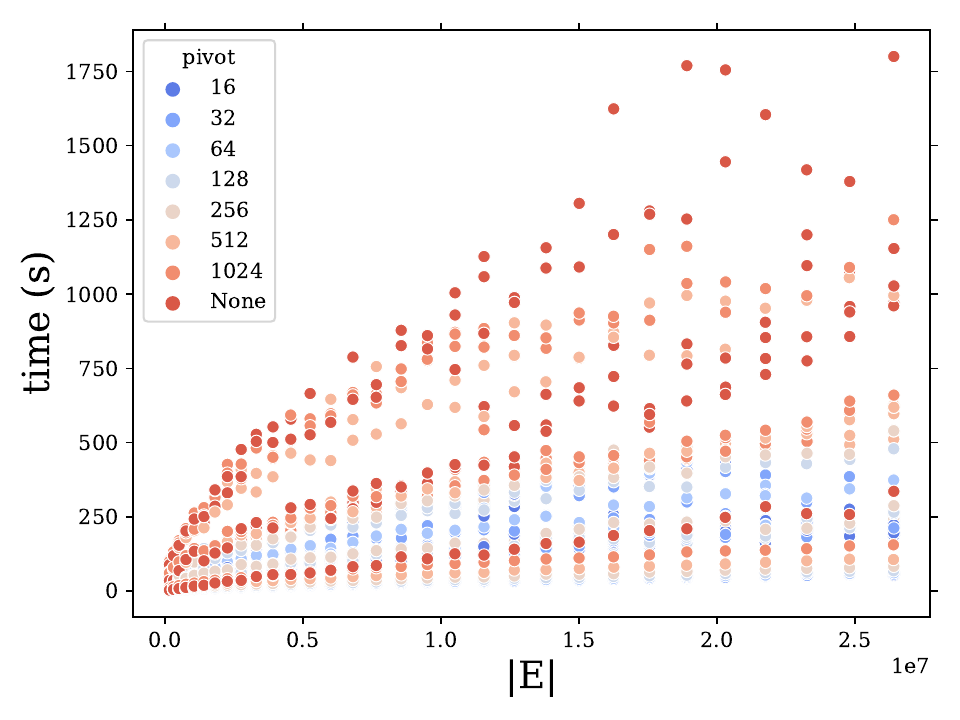}
        \caption{Time w.r.t. $|E|$.}
         \label{fig:scalability-abl2}
     \end{subfigure}
     \caption{Execution time with respect to the number of pivot nodes. Smaller number of pivot nodes induces substantially faster execution times.}
    \label{fig:execution-time}
\end{figure}

{We finally present the results on synthetic }{Erd\H{o}s-R\'enyi} {networks, where the effect of the number of pivot nodes on the execution time was studied. The main result is shown in Figure{~\ref{fig:execution-time}}. The result indicates that the number of pivot nodes can reduce the execution time by more than an order of magnitude -- with no pivoting nodes, DNR's execution time increases observably faster. More detailed results displaying the dependence with the node and link numbers are shown in subfigures} {~\ref{fig:scalability-abl}} and {~\ref{fig:scalability-abl2}}. The complexity, if {considering pivoting, remains linear with respect to the number of nodes (constant $d = p$ instead of $d = |N|$ in the symbolic step of DNR). Without pivoting, the complexity increases substantially, which is problematic for larger networks.}
{Consistent improvement with respect to the number of pivot nodes was observed -- the lower the pivot number, the faster the overall process. This observation confirms our theoretical analysis which indicated that substantial improvements could be observed, especially for larger networks. The results indicate that for larger networks comprised of tens of millions of edges, the pivoting-based solutions could offer more than an \emph{order of magnitude faster} embedding construction. For completion, tabular results summarised in this section are available as}~{\ref{appendix:tabular}}.

\subsection{Performance in a low-data regime}
\begin{figure}[htb]
    \centering
    \includegraphics[width = \linewidth]{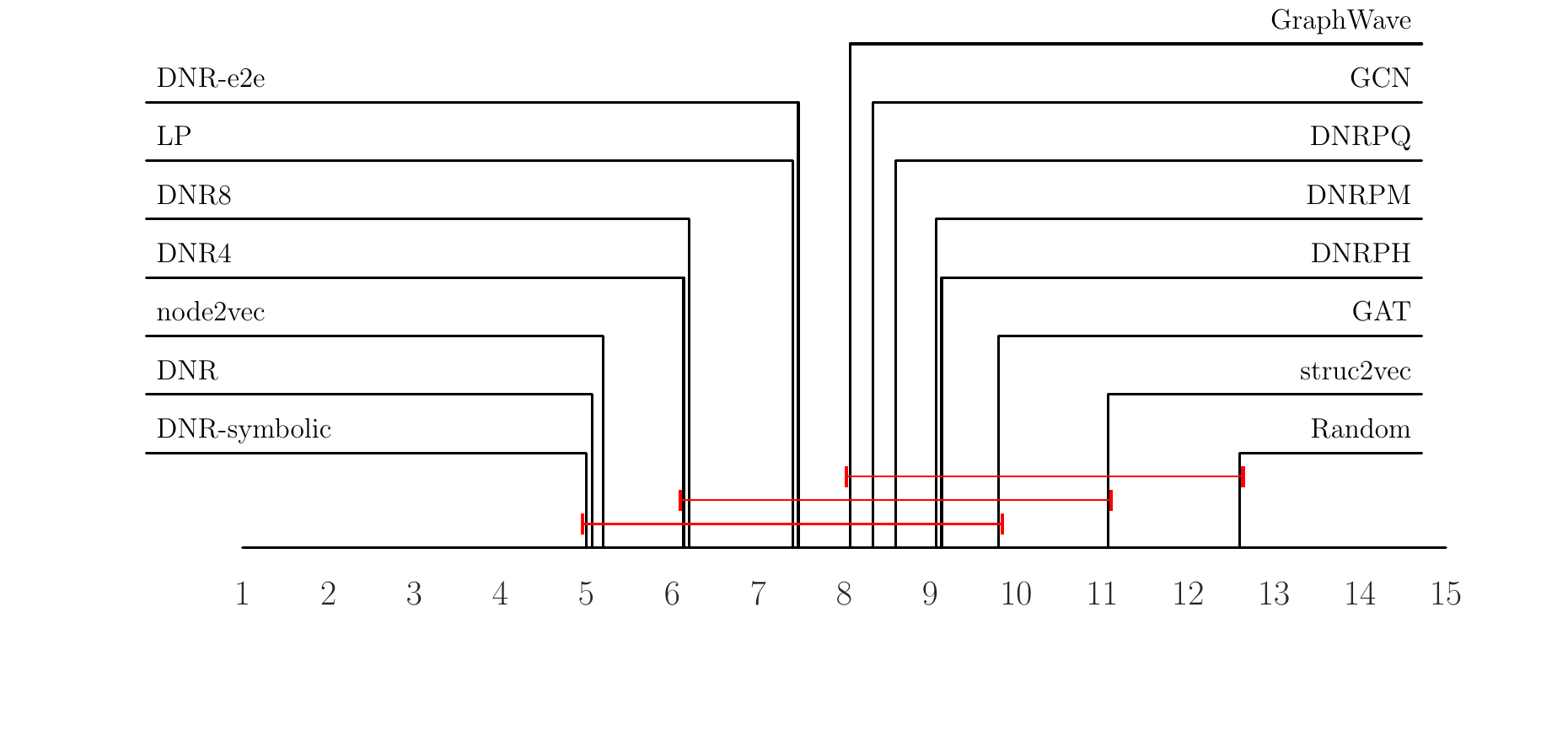}
    \caption{Micro F1 performance when the labelled data is scarce (10\%)}
    \label{fig:micro-minimal}
\end{figure}

{One of the main limitations of many existing node classification algorithms is their performance when only a small portion of a given network is labelled. We next present the DNR's behaviour when considering only 10\% of the labelled data in Figure}~\ref{fig:micro-minimal}.
{The experiment indicates that two of the DNR variants perform well when only a relatively few labelled data are available. This result potentially indicates the link between using a symbolic (global) rank space instead of using the more local, sampled walks, indicating that neuro-symbolic node representation learning has exciting potential for low-resource learning. Note that for larger data sets, this amount of labelled nodes can already be at the limit of what can be learned on a given commodity hardware setup, rendering DNR relevant for larger data sets.}

\subsection{Network visualization}

\begin{figure}[htb!]
    \centering
    \begin{tabular}{ccc}
\subcaptionbox{DNR.}{\includegraphics[width = .33\linewidth]{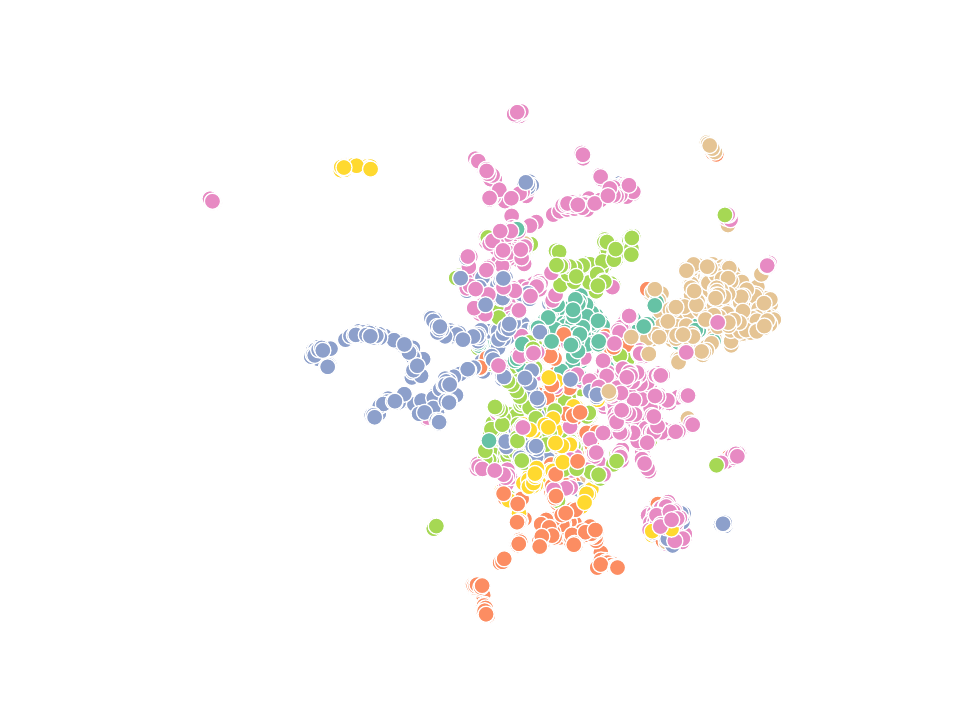}} &
\subcaptionbox{DNR (8 hidden).}{\includegraphics[width = .33\linewidth]{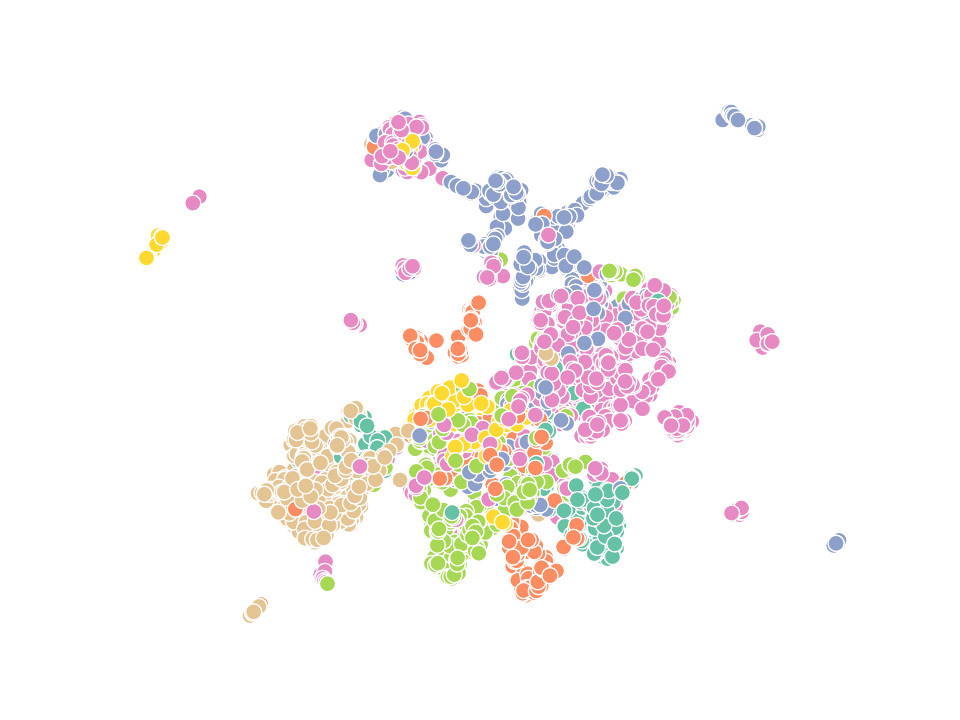}} & 
\subcaptionbox{Random embedding.}{\includegraphics[width = .33\linewidth]{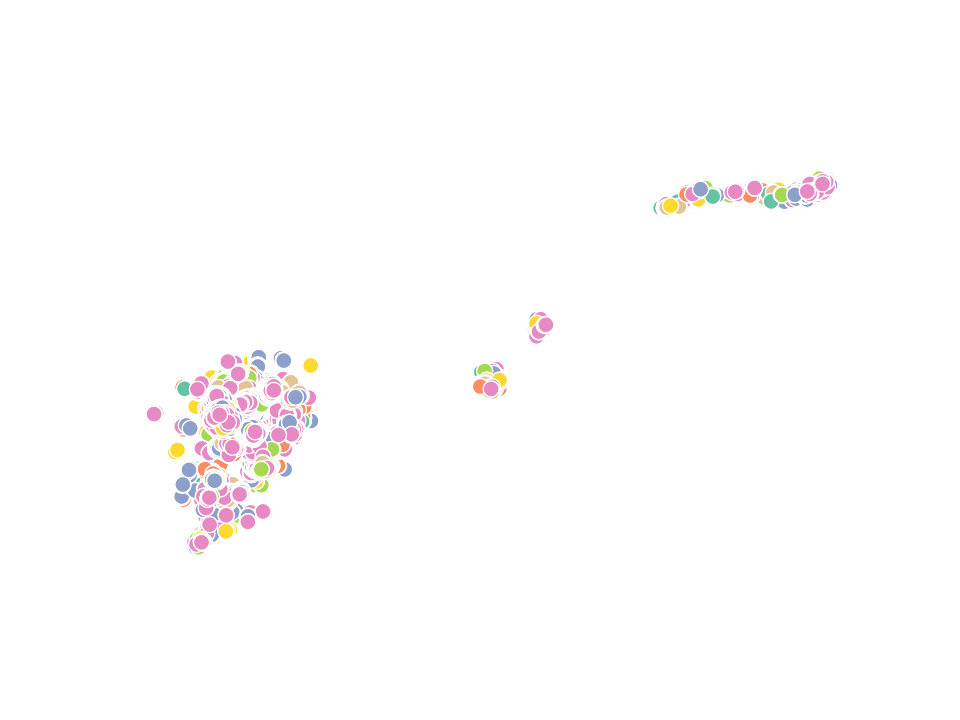}} \\
\end{tabular}
     \caption{Embedding visualization with overlaid labels (Cora). The random embedding is added as a reference of a non-structure-preserving projection visualization.}
    \label{fig:viz-emb}
\end{figure}

{We next demonstrate how DNR's results can be compressed to two dimensions with UMAP~{\cite{mcinnes2018umap}} to visualize a given network. By considering ten nearest neighbors with the minimum distance parameter set to 0.5, we obtained the visualization (colored by node labels) as shown in Figure~{\ref{fig:viz-emb}}.}

{The visualization shows distinct clustering patterns which to some extent also correspond to the label space (colors). Note that this representation was obtained in unsupervised manner, hence some variability with respect to label-position assignment is expected. A prominent use for this type of visualizations is when inspection of potentially interesting, structurally similar units is investigated via overlay of additional information. This visualization was, alongside the embedding to 2D, computed in a matter of seconds.}

\subsection{Comparing symbolic and subsymbolic representations}
\begin{figure}[htb!]
    \centering
    \begin{tabular}{ccc}
\subcaptionbox{DNR-symbolic ($d = |N|$).}{\includegraphics[width = .33\linewidth]{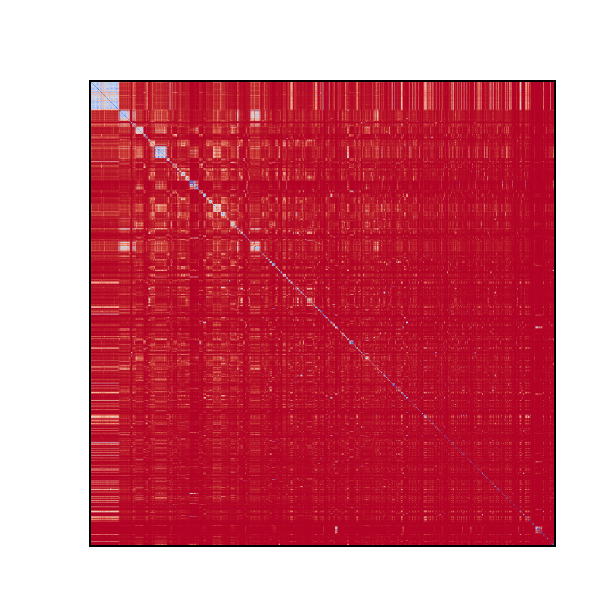}} &
\subcaptionbox{DNR ($d = 128$).}{\includegraphics[width = .33\linewidth]{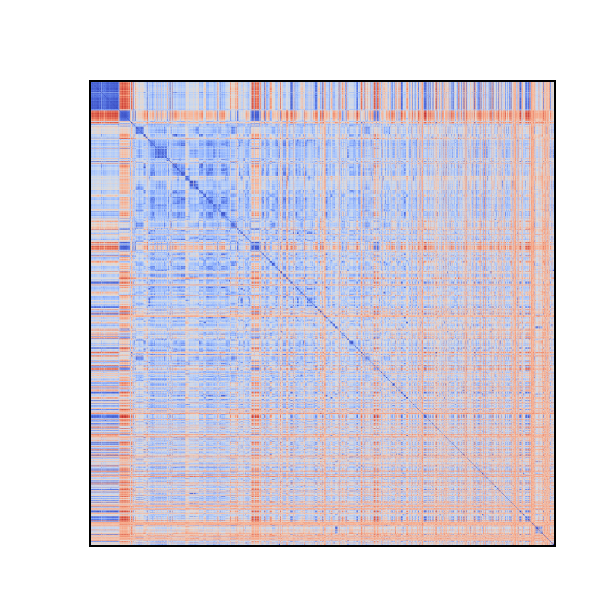}} & 
\subcaptionbox{$\textrm{abs}(\textrm{DNR-symbolic} - \textrm{DNR})$.}{\includegraphics[width = .33\linewidth]{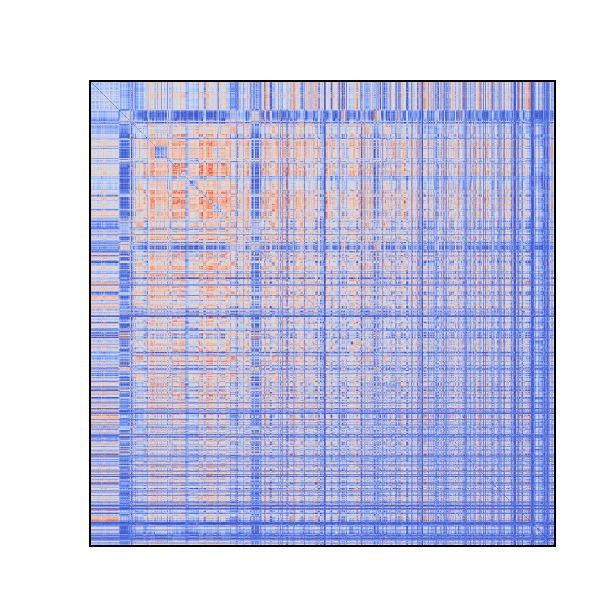}} \\
\end{tabular}
     \caption{{Representation space comparison. Each cell in (a) and (b) represents cosine distance between a given embedding pair. Cells in (c) represent the absolute difference values. The rows and columns correspond to the same nodes for all three sub-figures. Red represents high values and blue low ones.}}
    \label{fig:rep-comparison}
\end{figure}

{The proposed DNR's neuro-symbolic capabilities render it open for exploration of all intermediary representations and the relations between them. For the Cora network, we first computed node representations with DNR-symbolic ($d = |N|$) and DNR ($d = 128$) and investigated the distances between representations of individual nodes. For each node representation, we computed the cosine distance and normalized all values in the matrix by subtracting the minimum and dividing with the difference between the maximum and minimum to ensure a more fair comparison with respect to the distance bias for individual representations. The result is shown in }Figure~\ref{fig:rep-comparison}.
{We observe the following. The symbolic representation comparison matrix (a) mostly consists of entries indicating very high distances between a given embedding pair (intense red color $\implies$ higher distance). This observation indicates that representations in high dimensional spaces (in this case $d = |N|$) are far apart. The exceptions (similar nodes) are in the upper left part (blue). On the contrary, many more nodes are closer if we consider the more compact node representations obtained via the DNR algorithm (b). This result indicates that the neural network compresses the space (as expected), yielding fewer node representations that are distant from the others (red strips in the matrix). The final representation (c) represents the difference between the representations -- blue color in this case represents similar representations. The reader can observe that distant node representations obtained by DNR are relatively close to the ones obtained by DNR-symbolic (blue horizontal and vertical strips). The red squares in the upper left part, however, indicate node similarities that were not amplified by DNR-symbolic, but with DNR. This type of ablation is possible only for neuro-symbolic representation learners, and is to our knowledge one of the first of its kind for the considered task.}

\section{Discussion and conclusions}
\label{sec:conclusions}
In this work, we presented Deep Node Ranking, a methodology for scalable neuro-symbolic node embedding and direct end-to-end classification based on a given network's structure. In extensive empirical evaluation, we demonstrated DNR's competitive performance and superior scalability on multiple real-life and synthetic benchmark problems.

{The proposed methodology offers one of the first neuro-symbolic node representation learners -- the initial node features that are interpretable are compressed with a novel neural network architecture (DNRNet). Albeit the resulting representations are latent and non-interpretable to a human, the input to obtaining such a representation can be manipulated in a symbolic manner (e.g., effects of node removal), offering a simple-to-use testbed for investigating the effects of different structural interventions on a given network. Extensive empirical evaluation indicates that symbolic features are highly competitive. However, they could be impractical to compute, rendering the proposed neuro-symbolic variant of DNR highly useful for many contemporary network-based learning tasks. Furthermore, DNRNet performs well in low-data regimes, which was an interesting finding -- we expected that the symbolic-only variant would dominate in such settings.

We demonstrated that neuro-symbolic approaches could scale better than purely subsymbolic ones (e.g., node2vec or struc2vec), indicating that not all interpretability is necessarily sacrificed for good performance. We demonstrated that out-of-the-box DNR implementation performs competitively and in terms of micro F1 better than state-of-the-art, and further, it offers at least an order of magnitude speedup. By introducing the concept of \emph{node pivoting}, we demonstrate that DNR can scale to very large networks with tens of millions of links -- a scale where other considered methods do not operate well without specialized hardware. We confirmed the findings related to algorithms' performance with frequentist and Bayesian analysis. As Bayesian analysis was previously not conducted in such evaluation settings, we believe further work which will investigate the suitability (and scalability) of this branch of tests for network-related tasks is an interesting research direction.}

{In terms of further work related to the proposed algorithm, we see the following main directions. First, the effects of studying different pivoting schemes could offer better trade-offs between efficiency and performance. Next, by considering GPU-based implementations of the power iteration considered for computing the stationary random walk distributions, we believe additional speedups could be observed. Neuro-symbolic node ranking offers the direct study of the effects of perturbing specific, e.g., nodes and observing the properties of the resulting low-dimensional representations. Such \emph{structural interventions} potentially offer a more native explanation mechanism compared to \emph{post-hoc} approximation schemes considered in contemporary machine learning. Finally, we plan to explore the scalability of DNR across multiple machines by sharing the input network and performing the rankings only locally. Such implementation could scale to much larger networks than considered by current state-of-the-art approaches.}

{Finally, this work demonstrates that deeper neural networks are suitable models for structure-only learning, albeit, as shown, in a neuro-symbolic setting. Current results indicate that deeper neural networks are possible and potentially offer superior performance (unless overfitting takes place). A promising direction that would offer additional improvements is also the automatic development of neural network architectures via neuroevolution.}

\section{Availability}
\label{availability}
The DNR and the datasets allowed to be shared w.r.t. their licenses will be freely accessible at \url{https://github.com/SkBlaz/DNR}.

\subsection*{\textbf{Acknowledgments}}
The work of the first author was funded by the Slovenian Research Agency through a young researcher grant (B\v{S}).
The work of other authors was supported by the Slovenian Research Agency (ARRS) core research programs P2-0103 and P6-0411, 
and research projects J7-7303, L7-8269, and N2-0078 (financed under the ERC Complementary Scheme). The work was also supported by European Union's Horizon 2020 research and  innovation programme under grant agreement No 825153, project EMBEDDIA (Cross-Lingual Embeddings for
Less-Represented Languages in European News Media).

\bibliography{references.bib}
\appendix

\section{Tabular results with deviations}
\label{appendix:tabular}
In this section, we present the performance results for micro and macro F1 scores. The first table represents the micro F1 scores, followed by a table showing macro F1 scores. The runs marked with NaN reached the time-out point (did not finish).
\begin{table}[htb!]
    \centering
        \caption{Micro F1.}
        \resizebox{1\textwidth}{!}{
    \input{microtable}
    }
    \vspace{1cm}
            \caption{Macro F1.}
        \resizebox{1\textwidth}{!}{
    \input{macrotable}
    }
     \label{tab:tables}
\end{table}

\end{document}

%% file: microtable.tex
\begin{tabular}{llllllllllllllll}
\toprule
dataset &      Bitcoin & BitcoinAlpha &     Blogspot &  HomoSapiens &          POS &    chameleon &     citeseer &         cora &      cornell &  ecommerceTF &         film &         ions &     squirrel &        texas &    wisconsin \\
setting      &              &              &              &              &              &              &              &              &              &              &              &              &              &              &              \\
\midrule
DNR          &  0.71 (0.01) &  0.71 (0.01) &   0.2 (0.01) &   0.2 (0.01) &  0.45 (0.02) &  0.56 (0.04) &  0.55 (0.02) &  0.78 (0.02) &  0.52 (0.06) &   0.83 (0.0) &  0.34 (0.01) &  0.64 (0.03) &  0.41 (0.03) &  0.59 (0.07) &  0.53 (0.03) \\
DNR-e2e      &  0.69 (0.01) &   0.7 (0.01) &  0.09 (0.01) &  0.07 (0.01) &  0.41 (0.01) &  0.45 (0.04) &  0.57 (0.03) &  0.77 (0.04) &  0.49 (0.12) &  0.83 (0.01) &  0.35 (0.02) &  0.66 (0.03) &  0.39 (0.02) &  0.61 (0.06) &  0.49 (0.08) \\
DNR-symbolic &  0.72 (0.01) &  0.71 (0.01) &  0.28 (0.04) &  0.22 (0.03) &  0.45 (0.02) &  0.59 (0.04) &  0.62 (0.04) &   0.8 (0.04) &  0.54 (0.04) &  0.83 (0.01) &  0.36 (0.01) &  0.67 (0.04) &   0.5 (0.05) &  0.58 (0.07) &  0.49 (0.04) \\
DNR4         &  0.71 (0.01) &  0.71 (0.01) &  0.15 (0.02) &  0.15 (0.02) &    0.4 (0.0) &  0.52 (0.03) &  0.55 (0.01) &  0.74 (0.02) &  0.48 (0.05) &  0.83 (0.01) &  0.37 (0.01) &  0.63 (0.04) &  0.42 (0.02) &  0.58 (0.05) &  0.51 (0.03) \\
DNR8         &   0.7 (0.01) &   0.71 (0.0) &  0.14 (0.02) &  0.12 (0.01) &    0.4 (0.0) &   0.5 (0.02) &  0.54 (0.02) &  0.71 (0.03) &  0.53 (0.02) &  0.82 (0.01) &  0.37 (0.01) &  0.58 (0.05) &  0.41 (0.01) &  0.55 (0.04) &  0.52 (0.05) \\
DNRPH        &  0.68 (0.01) &  0.68 (0.01) &  0.15 (0.01) &  0.08 (0.01) &   0.4 (0.01) &  0.34 (0.01) &  0.26 (0.01) &  0.41 (0.01) &  0.54 (0.03) &    0.8 (0.0) &  0.34 (0.01) &  0.48 (0.01) &  0.39 (0.01) &  0.53 (0.03) &   0.5 (0.03) \\
DNRPM        &   0.69 (0.0) &    0.7 (0.0) &    0.1 (0.0) &  0.06 (0.01) &   0.4 (0.01) &  0.37 (0.01) &   0.2 (0.01) &   0.31 (0.0) &  0.52 (0.02) &   0.78 (0.0) &  0.37 (0.01) &  0.43 (0.01) &  0.41 (0.01) &  0.53 (0.02) &  0.51 (0.07) \\
DNRPQ        &  0.69 (0.01) &  0.69 (0.01) &   0.2 (0.01) &  0.11 (0.01) &  0.41 (0.01) &  0.42 (0.02) &   0.4 (0.02) &   0.6 (0.02) &  0.51 (0.03) &  0.83 (0.01) &  0.34 (0.02) &  0.58 (0.03) &  0.37 (0.01) &  0.54 (0.03) &  0.44 (0.06) \\
GAT          &   0.54 (0.0) &  0.55 (0.02) &          NaN &    0.0 (0.0) &  0.35 (0.12) &   0.6 (0.04) &  0.66 (0.07) &  0.82 (0.03) &  0.58 (0.05) &          NaN &   0.37 (0.0) &  0.51 (0.01) &          NaN &  0.65 (0.05) &  0.53 (0.03) \\
GCN          &  0.58 (0.02) &  0.57 (0.01) &  0.11 (0.01) &  0.03 (0.01) &  0.05 (0.05) &  0.61 (0.04) &  0.69 (0.05) &  0.83 (0.03) &  0.56 (0.04) &   0.82 (0.0) &  0.37 (0.01) &  0.19 (0.12) &  0.44 (0.01) &  0.63 (0.06) &  0.51 (0.02) \\
GraphWave    &   0.69 (0.0) &    0.7 (0.0) &    0.1 (0.0) &  0.07 (0.01) &   0.4 (0.01) &  0.37 (0.02) &  0.37 (0.01) &  0.42 (0.02) &   0.5 (0.03) &   0.78 (0.0) &  0.37 (0.01) &  0.53 (0.03) &  0.41 (0.01) &  0.54 (0.04) &  0.47 (0.04) \\
LP           &    0.7 (0.0) &  0.71 (0.01) &  0.22 (0.02) &   0.06 (0.0) &   0.07 (0.0) &   0.4 (0.02) &  0.64 (0.06) &  0.82 (0.04) &  0.54 (0.05) &  0.72 (0.02) &  0.35 (0.01) &  0.69 (0.06) &  0.41 (0.01) &  0.56 (0.02) &  0.42 (0.06) \\
Random       &  0.65 (0.02) &  0.66 (0.03) &  0.07 (0.01) &   0.06 (0.0) &  0.37 (0.03) &   0.3 (0.03) &   0.18 (0.0) &  0.21 (0.02) &  0.47 (0.06) &  0.78 (0.01) &  0.33 (0.02) &  0.36 (0.02) &  0.34 (0.04) &   0.5 (0.05) &  0.42 (0.03) \\
node2vec     &  0.69 (0.01) &  0.69 (0.02) &  0.36 (0.03) &  0.21 (0.02) &  0.52 (0.02) &  0.58 (0.02) &  0.58 (0.02) &  0.83 (0.03) &  0.51 (0.04) &  0.83 (0.01) &  0.34 (0.02) &  0.65 (0.03) &  0.43 (0.03) &  0.57 (0.05) &  0.51 (0.02) \\
struc2vec    &  0.67 (0.02) &  0.68 (0.01) &  0.08 (0.01) &   0.08 (0.0) &  0.38 (0.02) &  0.54 (0.03) &  0.27 (0.01) &   0.3 (0.02) &  0.45 (0.04) &  0.78 (0.01) &  0.34 (0.01) &  0.46 (0.02) &  0.41 (0.02) &  0.56 (0.04) &   0.53 (0.1) \\
\bottomrule
\end{tabular}

%% file: macrotable.tex
\begin{tabular}{llllllllllllllll}
\toprule
dataset &      Bitcoin & BitcoinAlpha &     Blogspot &  HomoSapiens &          POS &    chameleon &     citeseer &         cora &      cornell &  ecommerceTF &         film &         ions &     squirrel &        texas &    wisconsin \\
setting      &              &              &              &              &              &              &              &              &              &              &              &              &              &              &              \\
\midrule
DNR          &  0.32 (0.02) &   0.3 (0.01) &  0.06 (0.01) &  0.16 (0.02) &  0.06 (0.01) &  0.54 (0.04) &   0.5 (0.01) &  0.77 (0.02) &  0.25 (0.04) &  0.67 (0.01) &   0.2 (0.01) &  0.25 (0.05) &   0.3 (0.02) &  0.28 (0.06) &  0.39 (0.05) \\
DNR-e2e      &  0.27 (0.01) &  0.27 (0.01) &   0.01 (0.0) &  0.03 (0.01) &   0.04 (0.0) &  0.39 (0.07) &   0.5 (0.03) &  0.74 (0.06) &  0.22 (0.04) &  0.66 (0.02) &  0.17 (0.02) &   0.2 (0.02) &  0.18 (0.03) &  0.31 (0.04) &  0.32 (0.04) \\
DNR-symbolic &  0.31 (0.02) &  0.29 (0.01) &  0.11 (0.03) &  0.16 (0.04) &  0.06 (0.01) &  0.57 (0.04) &  0.56 (0.04) &  0.79 (0.05) &  0.21 (0.05) &  0.67 (0.04) &  0.19 (0.02) &  0.22 (0.05) &  0.39 (0.08) &  0.23 (0.06) &  0.27 (0.04) \\
DNR4         &  0.31 (0.02) &  0.28 (0.01) &  0.03 (0.01) &  0.09 (0.02) &   0.04 (0.0) &  0.48 (0.03) &  0.48 (0.01) &  0.72 (0.03) &  0.19 (0.02) &  0.65 (0.01) &  0.15 (0.01) &   0.2 (0.03) &  0.25 (0.03) &  0.26 (0.05) &  0.34 (0.04) \\
DNR8         &   0.3 (0.02) &  0.28 (0.01) &  0.03 (0.01) &  0.06 (0.01) &   0.04 (0.0) &  0.46 (0.02) &  0.47 (0.02) &  0.68 (0.05) &  0.19 (0.03) &  0.63 (0.01) &   0.14 (0.0) &  0.17 (0.02) &   0.2 (0.03) &  0.22 (0.04) &  0.26 (0.03) \\
DNRPH        &  0.29 (0.01) &  0.27 (0.01) &  0.06 (0.01) &  0.05 (0.01) &   0.04 (0.0) &  0.22 (0.01) &  0.21 (0.01) &  0.33 (0.02) &  0.26 (0.03) &  0.53 (0.01) &   0.2 (0.01) &  0.14 (0.02) &  0.24 (0.01) &  0.26 (0.04) &  0.33 (0.03) \\
DNRPM        &  0.27 (0.01) &  0.26 (0.01) &   0.01 (0.0) &   0.02 (0.0) &   0.03 (0.0) &   0.14 (0.0) &   0.06 (0.0) &   0.07 (0.0) &   0.17 (0.0) &   0.44 (0.0) &   0.14 (0.0) &   0.06 (0.0) &   0.15 (0.0) &  0.18 (0.02) &  0.17 (0.01) \\
DNRPQ        &   0.3 (0.01) &  0.28 (0.01) &  0.08 (0.01) &  0.09 (0.01) &   0.05 (0.0) &  0.37 (0.02) &  0.36 (0.02) &  0.58 (0.02) &  0.24 (0.03) &  0.66 (0.01) &  0.21 (0.01) &  0.22 (0.05) &   0.25 (0.0) &  0.28 (0.05) &  0.35 (0.08) \\
GAT          &   0.04 (0.0) &  0.05 (0.01) &          NaN &    0.0 (0.0) &   0.01 (0.0) &  0.59 (0.04) &  0.62 (0.07) &  0.81 (0.04) &  0.35 (0.16) &          NaN &  0.15 (0.01) &  0.08 (0.02) &          NaN &  0.36 (0.08) &  0.36 (0.05) \\
GCN          &  0.08 (0.01) &  0.08 (0.01) &   0.01 (0.0) &   0.01 (0.0) &  0.02 (0.01) &   0.6 (0.04) &  0.64 (0.05) &  0.82 (0.03) &  0.32 (0.13) &   0.64 (0.0) &  0.15 (0.01) &  0.08 (0.03) &  0.34 (0.01) &  0.35 (0.07) &  0.38 (0.05) \\
GraphWave    &  0.27 (0.01) &  0.26 (0.01) &    0.0 (0.0) &   0.03 (0.0) &   0.04 (0.0) &  0.21 (0.03) &  0.25 (0.01) &  0.21 (0.01) &  0.21 (0.02) &   0.44 (0.0) &   0.14 (0.0) &  0.14 (0.01) &   0.14 (0.0) &  0.18 (0.02) &  0.22 (0.03) \\
LP           &  0.28 (0.01) &  0.29 (0.01) &   0.1 (0.02) &   0.06 (0.0) &   0.06 (0.0) &  0.33 (0.03) &  0.61 (0.06) &  0.81 (0.04) &  0.24 (0.05) &  0.67 (0.02) &   0.2 (0.01) &  0.32 (0.04) &  0.24 (0.02) &  0.26 (0.02) &  0.28 (0.09) \\
Random       &  0.27 (0.01) &  0.26 (0.01) &   0.02 (0.0) &  0.05 (0.01) &   0.04 (0.0) &  0.25 (0.02) &   0.15 (0.0) &  0.13 (0.01) &  0.21 (0.03) &   0.44 (0.0) &  0.22 (0.01) &   0.08 (0.0) &  0.22 (0.01) &  0.23 (0.05) &  0.23 (0.02) \\
node2vec     &  0.33 (0.01) &   0.3 (0.01) &  0.23 (0.03) &  0.18 (0.02) &  0.11 (0.01) &  0.57 (0.02) &  0.54 (0.02) &  0.82 (0.03) &  0.23 (0.03) &   0.67 (0.0) &  0.21 (0.01) &  0.32 (0.05) &  0.36 (0.03) &  0.27 (0.04) &  0.37 (0.06) \\
struc2vec    &  0.29 (0.01) &  0.28 (0.01) &   0.03 (0.0) &  0.06 (0.01) &   0.06 (0.0) &  0.52 (0.03) &  0.24 (0.01) &  0.17 (0.01) &  0.21 (0.04) &   0.45 (0.0) &  0.21 (0.02) &  0.13 (0.01) &  0.34 (0.01) &  0.28 (0.07) &   0.3 (0.03) \\
\bottomrule
\end{tabular}